%% file: sn-main.tex
\documentclass[sn-mathphys]{sn-jnl}

\jyear{2021}%

\theoremstyle{thmstyleone}%
\theoremstyle{thmstyletwo}%

\theoremstyle{thmstylethree}%

\normalbaroutside 

\usepackage[capitalize,noabbrev]{cleveref}
\crefname{subsection}{Subsection}{Subsections}
\crefname{subsubsection}{Subsubsection}{Subsubsections}

\newcommand{\ie}{\emph{i.e.}}
\newcommand{\eg}{\emph{e.g.}}
\newcommand{\etal}{\emph{et al.}}
\usepackage{xcolor}

\begin{document}

\title[Solving morphological analogies: from retrieval to generation]{Solving morphological analogies: from retrieval to generation}


\author*[1]{\fnm{Esteban} \sur{Marquer}}\email{esteban.marquer@loria.fr}
\author*[1]{\fnm{Miguel} \sur{Couceiro}}\email{miguel.couceiro@loria.fr}

\affil[1]{\orgname{Université de Lorraine, CNRS, LORIA}, \orgaddress{\city{Nancy}, \postcode{F-54000}, \country{France}}}


\abstract{Analogical inference is a remarkable capability of human reasoning, and has been used to solve hard reasoning tasks. Analogy based reasoning (AR) has gained increasing interest from the artificial intelligence community and has shown its potential in multiple machine learning tasks such as classification, decision making and recommendation with competitive results. We propose a deep learning (DL) framework to address and tackle two key tasks in AR: analogy detection and solving. The framework is thoroughly tested on the Siganalogies dataset of morphological analogical proportions (APs) between words, and shown to outperform symbolic approaches in many languages.
Previous work have explored the behavior of the Analogy Neural Network for classification (ANNc) on analogy detection and of the Analogy Neural Network for retrieval (ANNr) on analogy solving by retrieval, as well as the potential of an autoencoder (AE) for analogy solving by generating the solution word.
In this article we summarize these findings and we extend them by combining ANNr and the AE embedding model, and checking the performance of ANNc as an retrieval method.
The combination of ANNr and AE outperforms the other approaches in almost all cases, and ANNc as a retrieval method achieves competitive or better performance than 3CosMul.
We conclude with general guidelines on using our framework to tackle APs with DL.}



\maketitle

\input{content.tex}

\backmatter

\bmhead{Funding}

This research was partially supported by the ANR project ``Analogies: from theory to tools and applications'' (AT2TA), ANR-22-CE23-0023,
by the TAILOR project funded by EU Horizon 2020 research and innovation program under GA No 952215,
and by the Inria Project Lab ``Hybrid Approaches for Interpretable AI'' (HyAIAI).

\bmhead{Code availability statement}
Our implementation is publicly available at GitHub (\url{https://github.com/EMarquer/morpho-analogy-amai}).
Code used to generate and manipulate the Siganalogies~\cite{siganalogies} V1 dataset is available at GitHub \url{https://github.com/EMarquer/siganalogies}.

\bmhead{Data availability statement}
The Siganalogies~\cite{siganalogies} V1 dataset (DOI: 10.12763/MLCFIE) is publicly available in the Dorel repository \url{https://dorel.univ-lorraine.fr/dataset.xhtml?persistentId=doi:10.12763/MLCFIE}.
Our trained models will appear in the Dorel repository \url{https://dorel.univ-lorraine.fr/dataset.xhtml?persistentId=doi:10.12763/I5ED78}, excluding the models reported in appendix due to their large volume.




\bmhead{Conflict of interest}
The authors declare that they have no conflict of interest.










\begin{appendices}

\input{appendix}





\end{appendices}


\bibliography{bib.bib}


\end{document}

%% file: content.tex
\section{Introduction}
Abstraction and analogical inference are remarkable capability of human reasoning~\cite{measure-inteligence:2019:cholet,analogy-making-cognition:2001:mitchel}. To some extent, analogical inference can be thought of as transferring knowledge from a source domain to a different, but somewhat similar, target domain by leveraging simultaneously on similarities and dissimilarities. 
Analogical inference has been used to solve hard reasoning tasks, and analogy based reasoning has gained increasing interest from the artificial intelligence community and has shown its potential with competitive results in multiple machine learning tasks such as classification, decision making and recommendation \cite{FahandarH18,FahandarH21,HugPRS19,analogy-making-ai:2021:mitchel}.
Furthermore, analogical inference can support data augmentation through analogical extension and extrapolation for model learning, especially in environments with few labeled examples \cite{ap-functions-boolean-examples:2017:couceiro}.
Also, it has been successfully applied to several classical natural language processing (NLP) tasks such as 
machine translation \cite{analogy-alea:2009:langlais},
several semantic~\cite{fam2016morphological,analogies-ml-perspective:2019:lim,analogies-ml:2021:lim} and morphological tasks \cite{minimal-complexity:2020:murena,detecting:2021:alsaidi,transfer:2021:alsaidi,analogy-overview-extended-preprint:2021:alsaidi,analogy-genmorpho:2022:chan,analogy-retrival:2022:marquer,transfer:2022:marquer},
 (visual) question answering~\cite{SadeghiZF15}, solving puzzles and scholastic aptitude tests  \cite{PeyreSLS19}, and target sense verification (TSV)~\cite{analogy-tsv:2022:zervakis} which tackles disambiguation.

Many analogy based reasoning approaches are built upon the notion of \textit{analogical proportions} (APs), statements of the form ``$A$ is to $B$ as $C$ is to $D$'' denoted $A:B::C:D$.
Significant effort has been done in the past three decades to formally define and describe the properties of APs, and we cover key approaches related to our work in \cref{sec:related-work}. 
In particular, two tasks are usually considered for APs: analogy detection and analogy solving. 
\textit{Analogy detection} refers to deciding whether a quadruple $A,B,C,D$ forms a valid analogy $A:B::C:D$, and \textit{analogy solving} is the task of finding possible values $x$ for which $A:B::C:x$ constitutes a valid analogy. Expressions $A:B::C:x$ for an indeterminate $x$ are referred to as \textit{analogical equations} or AP equations. We further detail the two tasks and related approaches to tackle them in \cref{sec:soa-clf,sec:soa-generation}.

Morphology has been used to develop analogy-based approaches \cite{copycat:1995:hofstadter-mitchel,complexity-hofstadter:2017:murena,tools-analogy:2018:fam-lepage,minimal-complexity:2020:murena} using various formal frameworks, further described in \cref{sec:related-work}.
Indeed, morphological APs can be seen as character-string APs \cite{complexity-hofstadter:2017:murena,minimal-complexity:2020:murena}, which can be generalized to APs between sequences of symbols and, in turn, to APs between structures. In that way, morphological APs relate strongly to many application settings of analogy. Additionally, morphological APs covers many regular phenomena of morphology and can be used to generate new but plausible words which are easily understandable by humans. This makes makes morphology an ideal empirical framework to develop and analyze approaches to tackle APs.

Morphological APs are also interesting as a standalone task.
Being able to successfully model morphological APs is useful to integrate morphological innovation in automatic systems and to generate linguistic resources, in particular for languages with complex morphology or that rely heavily on morphological innovation.
For instance, \cite{tools-analogy:2018:fam-lepage} proposes a toolkit to extract analogical clusters of word morphology, which summarize words with similar morphological behavior by relying only on text data.
This kind of result can serve as a basis for empirical analysis of morphology or as teaching material.
In that trend, our previous work \cite{transfer:2022:marquer} details results on the transfer of analogy detection models between languages, and showcases the use of morphological analogy to compare languages empirically.

All these aspects led us to develop a complete framework for AP manipulation using deep learning that we extensively tested on the morphology of more than 16 languages in the present article as well as other languages in our previous works~\cite{detecting:2021:alsaidi,transfer:2021:alsaidi,analogy-overview-extended-preprint:2021:alsaidi,analogy-genmorpho:2022:chan,analogy-retrival:2022:marquer,transfer:2022:marquer}.
This paper serves as a summary for previous work on analogy detection and analogy solving ~\cite{detecting:2021:alsaidi,transfer:2021:alsaidi,analogy-overview-extended-preprint:2021:alsaidi,analogy-genmorpho:2022:chan,analogy-retrival:2022:marquer,transfer:2022:marquer}, and extends on papers published in the ATA@ICCBR2022 and IARML@IJCAI2022 workshops~\cite{analogy-retrival:2022:marquer,analogy-genmorpho:2022:chan}. It also serves as a standalone reference for future used of the analogy detection and solving models.
In particular, we improve the results of the retrieval approach~\cite{analogy-retrival:2022:marquer} using word generation from~\cite{analogy-genmorpho:2022:chan}.
We compare the performance of all our models on analogy solving, including new model variants.

The paper is organized into two main parts.
\begin{itemize}
    \item In \cref{part:models} we recall the basic background on APs and we introduce the  main components of our framework. More precisely: 
\begin{itemize}
    \item We introduce the basic formalism of APs in \cref{sec:formal-framework} with a more detailed description of morphological APs in \cref{sec:morpho-ap}. We also present a brief summary of recent approaches in tackling APs with a focus on morphological APs in \Cref{sec:soa-clf,sec:soa-generation}.
    \item We describe the framework components in \cref{sec:models}, with the embedding models, analogy processing models, and training strategy described respectively in \cref{sec:emb,sec:model-ann,sec:training}.
\end{itemize}
    \item In \cref{part:expe}, we report experimental results on the morphological tasks considered in this paper, namely, state of the art results on multilanguage analogy detection and solving dealing with morphology. Specifically:
\begin{itemize}
    \item We present new experiments  performed on 16 languages of the Siganalogies~\cite{siganalogies} dataset described in \cref{sec:data,sec:data-split}, including languages not studied in our previous approaches~\cite{detecting:2021:alsaidi,transfer:2021:alsaidi,analogy-overview-extended-preprint:2021:alsaidi,analogy-genmorpho:2022:chan,analogy-retrival:2022:marquer,transfer:2022:marquer}. The list of all models used is detailed in \cref{sec:analogy-solving-expe}, and \cref{sec:analogy-solving-expe-results} covers the model performance.
    \item We also provide an overview of the results and conclusions from our previous and current experiments and provide guidelines for future use of the {ANNc}~/~{ANNr} framework (Analogy Neural Network for classification~/~for retrieval and generation) in \cref{sec:discussion}.
    Key aspects of the framework are discussed in details, such as the benefit of using ANNc and ANNr over other methods, the data augmentation process and its link with the axiomatic setting, the use of the framework in a multilingual context, as well as several limitations of the approaches used.
\end{itemize}
\end{itemize}
We conclude the paper with an overview on the main contributions together with  overall analysis of the framework in \cref{sec:ccl}.

\newpage
\part{Analogy modeling and solving}\label{part:models}
\section{Related approaches}\label{sec:related-work}
Analogy has been in a variety of settings and refers to similar but distinct notions.
Several efforts have been made to provide a unifying formal framework of analogy using different axiomatic, logical, model theoretic, and functional approaches~\cite{analogy-formal:2001:lepage,analogical-dissimilarity:2008:miclet,complexity-hofstadter:2017:murena,ap:2022:antic}. For instance, \cite{analogy-between-concepts:2019:barbot-etal}~separates analogies into: analogical proportions (APs), relational proportions, simile, and metaphor.
While there is not general consensus on the different notions of analogies, it is generally accepted that they can be reformulated as APs. This is true for instance for relational proportions from~\cite{analogy-between-concepts:2019:barbot-etal} that follow the same writing as APs, while simile and metaphor can be seen as APs where some elements are not explicitly expressed. 

In this paper, we focus on morphological APs, \ie, APs $A:B::C:D$ that capture morphological transformations of the four words $A,B,C,D$ (\eg, conjugation or declension). In \cref{sec:formal-framework} we introduce the formal framework we use for our approach, which follows the seminal work of~\cite{lepage1996saussurian} by exploiting the postulates of analogical proportions. We discuss the limitations of the formal framework in \cref{sec:discussion}.
To give the reader an overview of possible formulations of the problem of analogy, we also provide a brief introduction to other key formal frameworks.
In \cref{sec:morpho-ap} we expand on the notion of morphological transformation and the resulting morphological analogical proportions.
In \cref{sec:soa-clf,sec:soa-generation} we present key approaches to detecting and solving morphological analogical proportions.
As deep learning approaches to morphological analogies are strongly related to approaches on semantic word analogies, the latter will also be discussed here.

\subsection{Analogical proportions and other formal frameworks of analogy}\label{sec:formal-framework}
In this work we focus on the notion of analogical proportions (APs)~\cite{trends-analogical-reasoning:2014:prade,analogical-dissimilarity:2008:miclet}, and in particular we follow the axiomatic setting introduced by Lepage~\cite{analogy-commutation-linguistic-fr:2003:lepage} consisting in 4 axioms in the linguistic context for analogical proportions: \textit{symmetry} (if $A:B::C:D$, then $C:D::A:B$), \textit{central permutation}  (if $A:B::C:D$, then $A:C::B:D$), \textit{strong inner reflexivity} (if $A:A::C:D$, then $D = C$), and \textit{strong reflexivity} (if $A:B::A:D$, then $D = B$).
These postulates imply several other properties, for instance \textit{identity} ($A:A::B:B$ is always true), \textit{inside pair reversing} (if $A:B::C:D$, then $B:A::D:C$) and \textit{extreme permutation} (if $A:B::C:D$, then $D:B::C:A$).
This axiomatic setting is related to the common view of APs as geometric (\cref{eq:geom-proportion}) or arithmetic proportions (\cref{eq:arith-proportion}) or, in geometric terms, as parallelograms in a vector space (\cref{eq:parallelogram}):
 \begin{equation}
  \frac{A}{B}=\frac{C}{D},
  \label{eq:geom-proportion}
 \end{equation}
 \begin{equation}
  A-B=C-D,
  \label{eq:arith-proportion}
 \end{equation}
\begin{equation}
    \vec{A}-\vec{B}=\vec{C}-\vec{D}
    \label{eq:parallelogram}.
\end{equation}
Most of the works on analogy relying on such a formalization manipulate APs between symbols and strings of symbols ($abc:abd::efg:efh$), which can be directly related to morphology ($cat:cats::dog:dogs$), but it is also possible to apply the axiomatic setting to other kinds of analogy for instance Boolean data as was done in \cite{boolean-analogy:2018:couceiro,galois-analogical-clf:2023:couceiro-lehtonen}.
While these axioms seem reasonable in the word domain, they can be criticized in other application domains~\cite{ap:2022:antic}.

In~\cite{transfer:2022:marquer}, we explore some limitations of the axiomatic setting for training models of morphological APs, and in particular how accepting or refusing central permutation impacts the performance of our ANNc models.

\subsection{Morphological analogical proportions}\label{sec:morpho-ap}
In this paper, we focus on the study of morphological APs, that is, APs between words in which the underlying relation is a morphological transformation.
A morphological transformation is a transformation of the structure of a word that follows a set of morphological rules, defined by the language. 

Morphology is usually separated into \textit{inflectional morphology} and \textit{derivational morphology}.
On the one hand, derivational morphology refers to morphological transformations that allow to create new words in a systematic manner, for instance in English the prefix \textit{un-} allows to create \textit{unaware} from \textit{aware} in the same manner as \textit{untold} from \textit{told}.
On the other hand, inflectional morphology describes morphological transformations that express a change in the grammatical nature of a word without altering the core meaning of the word. For example, \textit{looked} is the result of a change of \textit{tense} in the English verb \textit{to look} by adding the suffix \textit{-ed}.
The data we use in our approach is extracted from inflectional morphology datasets as described in \cref{sec:data}, so our experiments are limited to APs on inflectional morphology. 
Given that the morphological mechanisms of derivational and inflectional morphology are very close, this limitation in the scope of our work could easily be solved by using data containing derivational morphology, without modifying our approach.
Additionally, examples presented in this paper contain both derivational and inflexional morphological transformations, to better illustrate the mechanisms of morphology.

\subsection{Analogy detection}\label{sec:soa-clf}
Analogy detection can be seen as a classification task: given a quadruple $A, B, C, D $, we classify the AP $A:B::C:D$ as valid or invalid according to our notion of AP.
It can be applied to extract analogies from data, for instance the tools in~\cite{tools-analogy:2018:fam-lepage} are designed to generate analogical grids, \ie, matrices of transformations of various words, similar to paradigm tables in linguistics~\cite{fam2016morphological}.
Each row of an analogical grid contain words using the same root, and going from one column to another corresponds to a single morphological transformation no matter the row.
By taking two rows and two columns of an analogical grids, we obtain an AP.
This approach~\cite{tools-analogy:2018:fam-lepage} detects morphological analogies using manually designed features such as the number of character occurrences and the length of the longest common subword.

Analogy detection on a single quadruple is less trivial than it may appear, as in many cases the boundary between valid and invalid analogies is not clearly defined.
A solution is to use machine learning to learn the boundary from data, as Lim \etal{} implemented~\cite{analogies-ml-perspective:2019:lim} for semantic word analogies and further explored in \cite{analogies-ml:2021:lim}.
Using a dataset of semantic APs, they learn an artificial neural network to classify quadruples $A,B,C,D$ into valid or invalid analogies, using pretrained word embeddings $e_A$, $e_B$, $e_C$, and $e_D$.
We follow a similar approach to morphological analogies in \cite{detecting:2021:alsaidi} by replacing the GloVe \cite{glove:2014:pennington} semantic embeddings used in~\cite{analogies-ml-perspective:2019:lim,analogies-ml:2021:lim} with a morphology-oriented word embedding model. The classifier is detailed in \cref{sec:model-annc-clf,fig:model-annc} under the name Analogy Neural Network for classification (ANNc).

\subsection{Analogy solving by retrieval and by generation}\label{sec:soa-generation}
Analogy solving is the process of completing an incomplete analogy, in particular AP equations (or analogical equations) like $A:B::C:x$ where $x$ is unknown.
Analogy solving strongly relates to the concept of analogical innovation, an important mechanism of creativity~\cite{analogy-making-cognition:2001:mitchel,analogy-making-ai:2021:mitchel}.
It can also be seen as a transfer operation, where $A$ is a source situation with a solution $B$, and $C$ is a target situation for which we want the solution.
The notion of analogy is particularly useful in this setting as it leverages simultaneously the similarities and differences between $A,B,C$ to adapt $B$ to the target situation $C$ and obtain the solution.

The \emph{Alea} algorithm~\cite{analogy-alea:2009:langlais} relies on the number of character occurrence in $A,B,C$ to determine the characters occurring in $D$ the solution to the analogical, and proposes a Monte-Carlo estimation of the permutation of these characters to solve the AP equation.
More specifically, \cite{analogy-alea:2009:langlais} follows the results of \cite{finite-state-trancducers:2003:yvon} about closed form solutions and proposes a Monte-Carlo estimation of the solutions of an analogical equation by sampling among multiple sub-transformations. 
With $bag(A)$ being the set of all characters in $A$, \emph{Alea} considers $bag(D)=(bag(B) - bag(A)) + bag(C)$ and thus $D$ is a permutation of the characters of $bag(D)$.
For example, the solution of $cat:cats::animal:x,x=animals$ is a permutations of the characters $\{a,a,i,l,m,n,s\}$, which can be obtained from the characters in $cats$ but not in $cat$ (\ie, $\{s\}$) together with the characters of $animal$ (\ie, $\{a,a,i,l,m,n\}$). This process cannot handle mechanisms like reduplication (repeating part of a word) that requires more complex comparisons between character occurrences.

A similar approach was later proposed in \cite{sigmorphon:2017:lepage}, using the postulates of \cite{lepage1996saussurian} to address multiple characteristics of words, such as their length, the occurrence of characters and of patterns. Based on these features, the author proposes an algorithm to solve analogies between character strings by first extracting relevant features, then estimating the position of each character using an arithmetic on character positions, and finally generating the corresponding word. The method accounts for multiple potential solutions by comparing the predicted word with permutations of its characters.

A more empirical approach which does not rely on the axioms of APs was proposed by \cite{minimal-complexity:2020:murena}, and considers some transformation $f$ such that $B=f(A)$ and $f(C)$ is computable.
Following the observation that humans tend to use the simplest thinkable analogy to solve analogical equations~\cite{complexity-hofstadter:2017:murena,minimal-complexity:2020:murena}, the simplest transformation $f$ is found by minimizing its Kolmogorov complexity.
This is done by generating $f$ by combining simple operations (insertion, deletion, \textit{etc.}) and computing the length of the resulting program.
With this process, not only obtain the solution of the AP equation but also the transformation $f$ is obtained.
Additionally, \textit{Kolmo} is able to handle more complex mechanisms than \emph{Alea} including reduplication.

In~\cite{analogies-ml-perspective:2019:lim,analogies-ml:2021:lim}, a retrieval approach was proposed for semantic APs and later adapted to morphological APs in~\cite{analogy-overview-extended-preprint:2021:alsaidi,analogy-retrival:2022:marquer}.
Similarly to the analogy detection approach proposed by the authors (see \cref{sec:soa-clf}), the analogy solving approach relies on pre-trained embeddings. An artificial neural network is used to predict the embedding of a solution, and the corresponding word is retrieved from a list of possible words based on its embedding.
A similar neural network is detailed in \cref{sec:model-annr,fig:model-annr} under the name Analogy Neural Network for retrieval/generation (ANNr).

Recently, \cite{vec-to-sec-sent-analogies:2020:lepage} proposed a generation framework to solving sentence analogies.
They use an autoencoder model (named ConRNN) trained to reconstruct sentences, and perform simple arithmetic operations on the embedding space to solve analogies. Once the analogy between embeddings is solved, the decoder part of ConRNN is used to generate the solution from the predicted embedding.
Their model is a sequence-to-sequence model composed of 2 elements. First, a sentence (as a sequence of words) is used as input to an encoder RNN, and the last hidden state of the RNN is used as the sentence embedding.
The latter is then fed to a decoder RNN that tries to predict the words of the input sentence.
The use of a generative model achieves significantly better results than the nearest neighbor algorithm on the same embedding space.
With a similar sequence-to-sequence autoencoder (AE) model as \cite{vec-to-sec-sent-analogies:2020:lepage}, we proposed in \cite{analogy-genmorpho:2022:chan} an approach to generate words at the character level, and using simple vector arithmetic we achieved significant performance improvement over baselines on solving morphological AP equations.
One of the aims of this paper is to further improve the performance by integrating the ANNr approach developed in \cite{analogies-ml-perspective:2019:lim,analogies-ml:2021:lim,analogy-retrival:2022:marquer} with the AE model.

\section{Overview of the framework}\label{sec:models}
\begin{figure}[h]
    \centering
    \includegraphics[width=\textwidth]{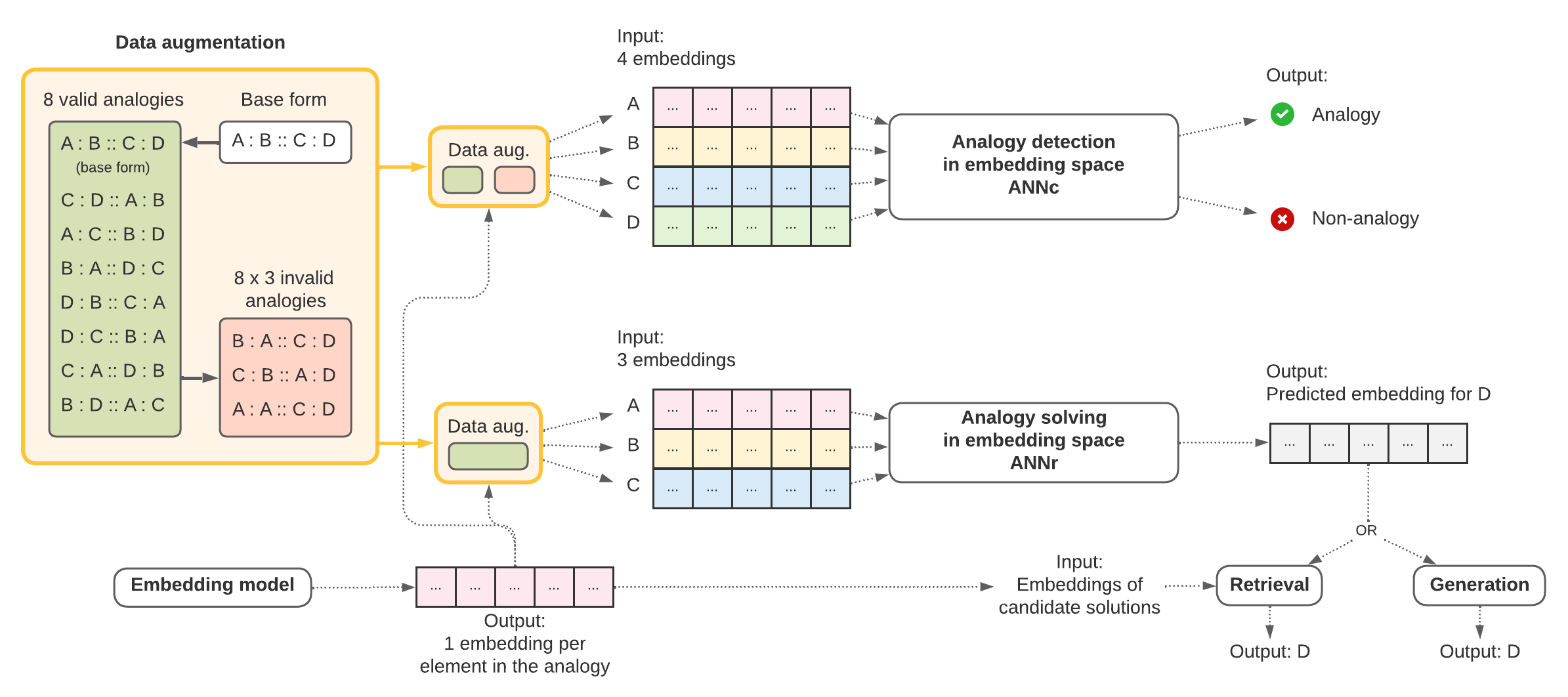}
        \caption{Overview of the framework: morphological embedding models, data augmentation, analogy classification (ANNc) and analogy solving (ANNr) models.}
    \label{fig:model-summary}
\end{figure}

Our framework can be split into two groups of models, namely the embedding models and the analogy models, summarized in \cref{fig:model-summary}.

To obtain suitable representations of words, we use two different kinds of embedding models described in \cref{sec:emb}: an embedding model designed for morphological applications and inspired from \cite{morpho-thesis:2020:vania} and an autoencoder (AE) model following a traditional sequence to sequence approach.

To detect and solve analogies, we rely on the two neural network models described in \cref{sec:model-ann}: the \textit{Analogy Neural Network for classification} (ANNc) and \textit{Analogy Neural Network for retrieval/generation} (ANNr), which were initially introduced in \cite{analogies-ml-perspective:2019:lim} and refined along our experiments.
The architecture of ANNc and ANNr is based on intuitions driven by the axiomatic setting as described in \cref{sec:formal-framework}.
In addition to these, we experiment with multiple non-parametric approaches such as the \textit{parallelogram rule} and \textit{3CosMul}~\cite{3-cos-mul:2014:levy-goldberg} that we will also describe in \cref{sec:model-ann}.

We separate retrieval approaches from generation approaches, as the open world assumption is used for the generation model and the closed world assumption for the various retrieval models.
The closed world assumption means that we know all the solution candidates, and no solution outside of these is accepted. Conversely, in the open world assumption the space of solution is not known beforehand, and we can obtain solutions beyond what is present in the data.

The data augmentation described in \cref{sec:training-annc,sec:training-annr} enables training models to fit the formal postulates of AP (see \cref{sec:formal-framework}) by becoming invariant to them.

\section{Embedding models}\label{sec:emb}
An important part of many machine learning systems is finding features relevant to the task to accomplish and using the features to represent the data. 
In deep learning, this task is usually handled by a family of approaches called \textit{embeddings}, real-valued vector representations of the data obtained by learning an \textit{embedding model} from data.
It is commonly accepted that the quality of such representations, which corresponds to the amount and nature of the information contained in the embedding as well as the properties of the embedding space, is a key factor in the performance of deep learning approaches.
Learning an embedding model of high quality often requires high amounts of data and significant training time which can be challenging, so for many applications on text pre-trained large-scale embedding models are made available, for instance, Bert \cite{bert-multilingual:2019:delvin}.
However, in our early experiments with pretrained word embeddings models (in particular, GloVe~\cite{glove:2014:pennington}, {word2vec}~\cite{efficient-representation-w2v:2013:mikolov}, and {fasttext}~\cite{fasttext:2017:bojanwoski}) we obtained poor performance\footnote{
We reproduced a similar experiment on the English data in \cref{app:sem-emb}.
}, which can be easily understood as these models contain information of a semantic nature\footnote{To be exact, these models relate to distributional semantics, which is based on the co-occurrence of words in text.} while we need morphological information to deal with morphological APs.

To solve this issue, we used with two different embedding models: a model inspired from \cite{morpho-thesis:2020:vania}, using a \textit{convolutional neural network (CNN)}\cite{cnn:1989:lecun}, and a model based on the autoencoder (AE) technique~\cite{ae:1991:kramer} and \textit{Long- and Short-Term Memory network (LSTM)}\cite{lstm:1996:hochreiter}.

\subsection{CNN-based embedding model}
\begin{figure}[h]
    \centering
    \includegraphics[width=.65\textwidth]{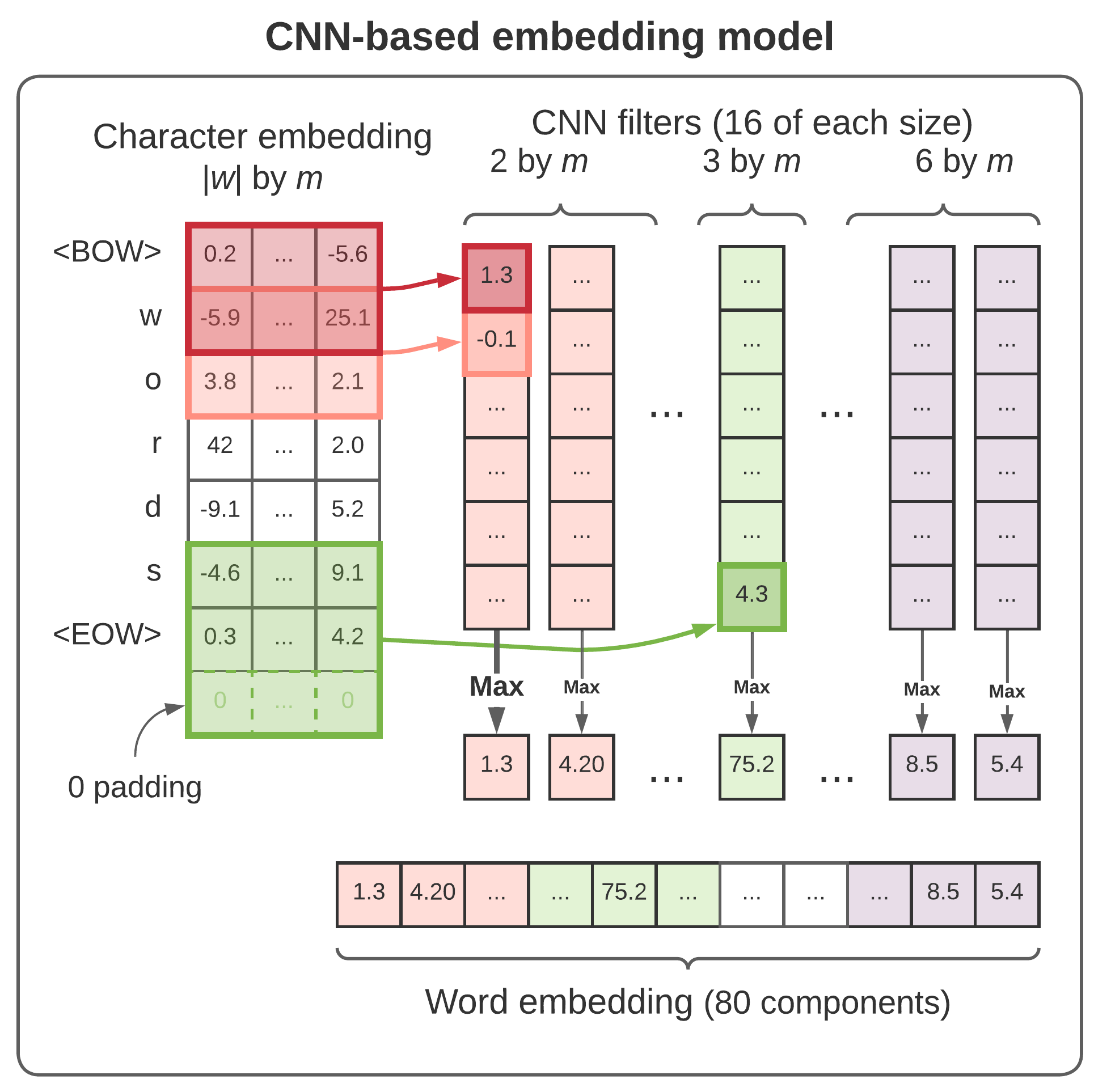}
        \caption{Morphological word embedding model based on character-level CNNs. The special characters \texttt{<BOW>} and \texttt{<EOW>} allow the CNN filters to identify characters at the beginning and end of the word.}
    \label{fig:model-cnn}
\end{figure}
The CNN-based model learns to detect key morphological patterns from the characters forming a word.
To do so, the input of the model are the characters of a word.
First, the characters are embedded into vectors of size $m$ learned together with the rest of the word embedding model. 
As schematized in \cref{fig:model-cnn}, multiple CNN filters are used, and each filter goes over the character embeddings by spanning over the full embeddings of $2$ to $6$ characters, resulting in filter sizes between $2$ by $m$ and $6$ by $m$.
For each filter, the model computes the maximum output to serve as a component of the word embedding. We use $16$ filters of each size between $2$ to $6$, resulting in embedding of $80$ components.
This last operation keeps only salient patterns detected by each of the filters, and forces each CNN filter to specialize in identifying a specific pattern of characters.
By using character embeddings to encode character features, the model is able to capture patterns based on character such as ``-ing'' but also based on features of characters like ``vowel-vowel-consonant''.
This flexibility is useful to deal with phenomena like \textit{euphony} which is, in very simple terms, a change of sound to make a word easier to pronounce (\eg, \textit{far} becomes \textit{further} when adding the suffix \textit{-ther}).
These character patterns correspond to \textit{morphemes}, which are the minimal units of morphology.
As the main components of this embedding model are CNN filters, we coin it the CNN-based model.

When used for analogy solving, the CNN-based model does not allow us to use an arbitrary embedding to generate the solution.
Instead, we need to compute the embeddings of a list of candidate words and select the most relevant word accordingly, making the space of solution closed.
This property is not an issue when working in a closed world assumption (\textit{i.e.}, we know all the solution candidates, and no solution outside of these is accepted), but in an open world assumption, we would theoretically have to compute the embeddings of all possible character strings, which is not feasible.

\subsection{AE embedding model}
\begin{figure}[h]
    \centering
    \includegraphics[width=.7\textwidth]{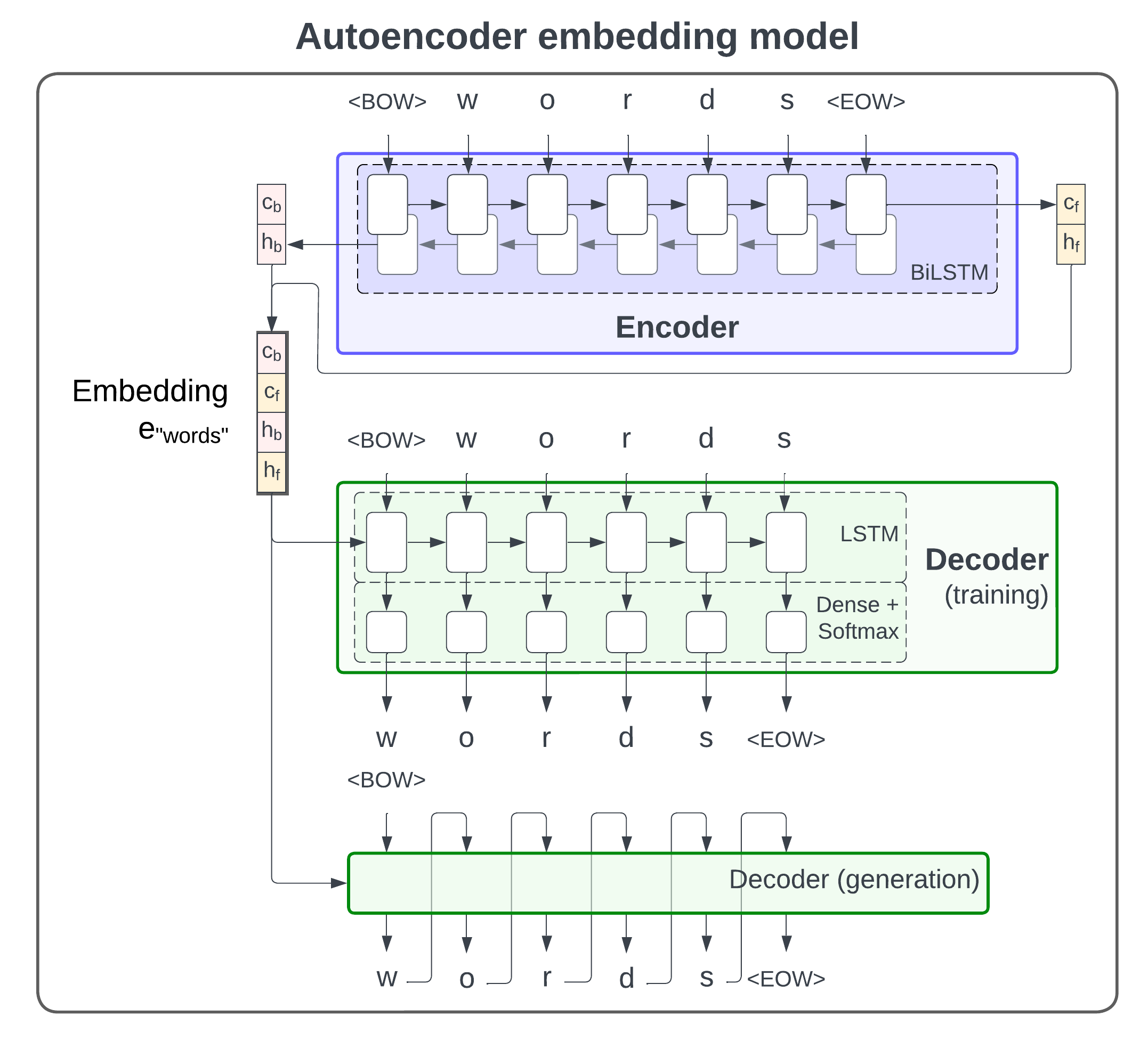}
        \caption{Morphological word embedding model based on character-level CNNs. The special characters \texttt{<BOW>} and \texttt{<EOW>} mark the beginning and end of the word, and are used in the generation process.}
    \label{fig:model-ae}
\end{figure}
Our AE model stems from the above-mentioned limitation of the CNN-based model for analogy solving.
To tackle this issue, we use an AE model that learns the embedding of a word from its characters and learns to generate word from embeddings.

An AE can be seen as a lossy compression algorithm, composed of an \textit{encoder} and a \textit{decoder}.
Taking an example in our setting, the encoder compresses the information of a word into an embedding and the decoder decompresses the embedding back into a word.
In order to properly decode the embedding (and to minimize the compression loss in our compression algorithm analogy), the model is trained to encode words and then decode the resulting embedding back into the original word. The difference between the original and the decoded word is used as the training objective in the \textit{autoencoding task}.
Because the model is able to recreate the original word from the embedding, the latter contains the key information to represent the former.
In more details, the decoder will learn to reproduce systematic or redundant parts of the training data without relying on the embedding, while key information to differentiate the training are encoded in the embedding by the encoder.

The architecture for our model is a character-level sequence-to-sequence autoencoder model, based on the model described in \cite{fchollet-seq2seq}.
We use Long- and Short-Term Memory network (LSTM) \cite{lstm:1996:hochreiter}, which are neural networks designed to handle sequences, and which allow us to encode and decode embeddings of a constant size no matter the length of the word, which we see as a sequence of characters.
Indeed, in our setting we need morphological information, which requires knowing the characters composing a word.
Each character of a word is encoded into a one-hot vector\footnote{A one-hot encoding vector of an element $x_i$ for a set $\{x_1,\dots,x_n\}$ is a vector of size $n$ containing $0$ for all components except component $i$ which is $1$.} and fed into the encoder, which is a Bidirectional LSTM (BiLSTM)\cite{bilstm:2005:graves}, a variant of the LSTM that reads the elements of the input sequence in the forward and backward order simultaneously.
This layer outputs four vectors: the last hidden state $h_f$ and cell state $c_f$ in the forward direction, and similarly $h_b$ and $c_b$ for the backward direction.
The concatenation of these vectors $e_w = \text{concat}(h_f,h_b,c_f,c_b)$ is the embedding of the word.
The decoder is an LSTM followed by a fully connected layer (or perceptron or dense layer) with softmax activation.
The input for the first step of the decoder is the above-mentioned embedding, split into two states $h=\text{concat}(h_f,h_b)$ and $c=\text{concat}(c_f,c_b)$.
The output of the decoder is a sequence of character predictions, with each prediction being a probability distribution over all the characters seen in training obtained with the softmax activation.

The process of decoding is usually a step by step process where each character of the output is predicted one after the other, with the previous character used as input to predict the next one. The first character is primed with  \textit{beginning of word} (BOW) character that marks the start of the word, and prediction is stopped when we encounter an \textit{end of word} (EOW) character.
During training, we accelerate training by using \textit{teacher forcing}, a variant of this decoding process where the characters of word to predict are used in place of the predicted characters as input for the next steps as illustrated in \cref{fig:model-ae}.

\section{Models for analogy processing}\label{sec:model-ann}
To manipulate APs, we propose multiple neural network models, building upon the embedding models.
Here we describe the structure and inner workings of these models, as well as how to use them for analogy detection and analogy solving.
We also describe three approaches that do not rely on neural networks to solve AP equations, namely, 3CosAdd~\cite{efficient-representation-w2v:2013:mikolov} and 3CosMul~\cite{3-cos-mul:2014:levy-goldberg}, and the parallelogram rule. 

\subsection{Analogy detection using ANNc}\label{sec:model-annc-clf}
\begin{figure}[h]
    \centering
    \includegraphics[width=.9\textwidth]{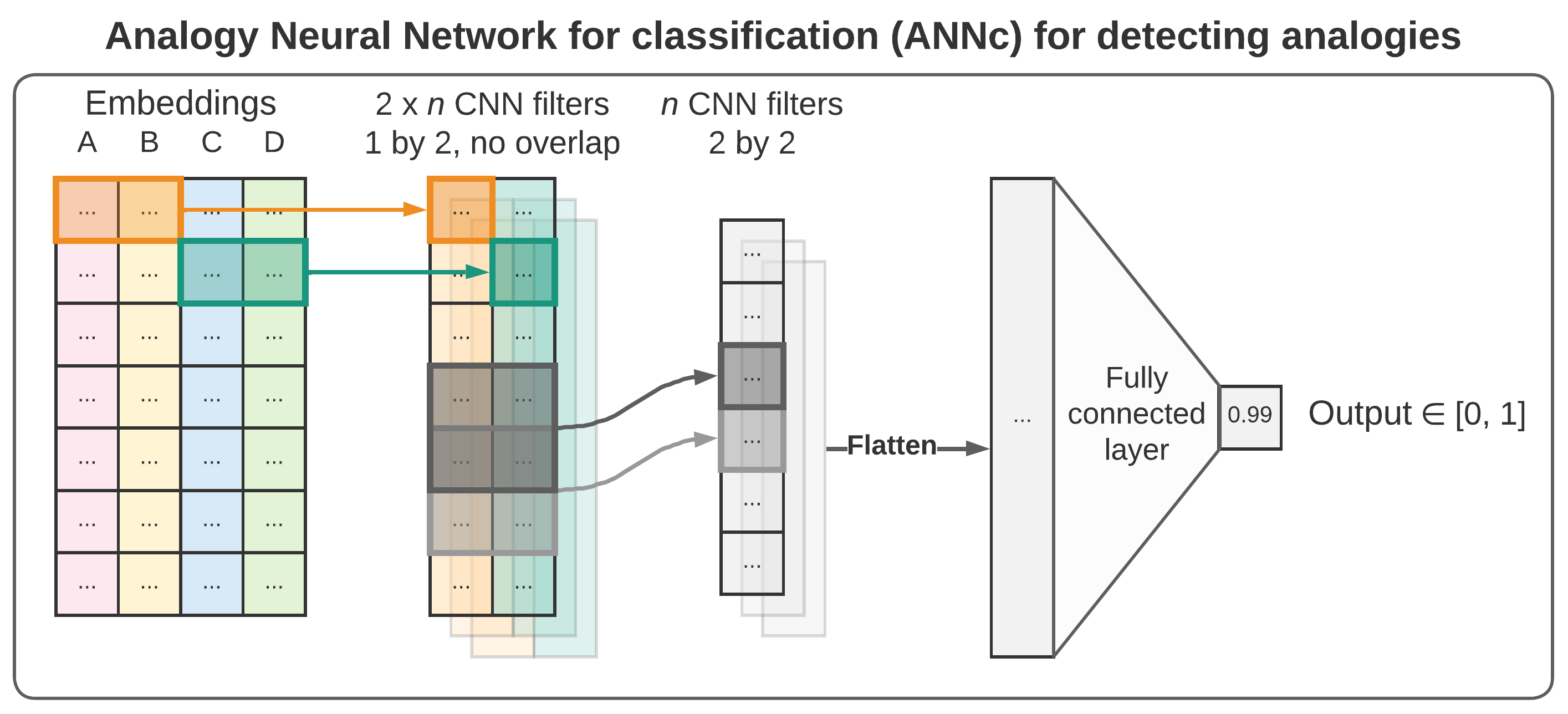}
        \caption{Analogy Neural Network for classification (ANNc). The embedding of the four input elements $A,B,C,D$ are shown vertically, each in a different color.}
    \label{fig:model-annc}
\end{figure}
The {Analogy Neural Network for classification (ANNc)} follows the idea that a quadruple $A,B,C,D$ constitutes a valid analogy $A:B::C:D$ if $A$ and $B$ differ in the same way as $C$ and $D$.
If this is true for all the features of $A,B,C,D$, \textit{i.e.}, for all dimensions of the embeddings $e_A,e_B,e_C,e_D$, then the analogy is true.

The model we use is based on multiple CNN layers taking as input $e_A,e_B,e_C,e_D$, each of size $n$, stacked into a $n\times4$ matrix.
Then, a first set $F_1$ of CNN filters of size $1\times2$ is applied on the embeddings, such that for each component $\cdot_i$ of the embedding vector, each filter spans across $A_i$ and $B_i$ simultaneously, and  across $C_i$ and $D_i$ simultaneously, with no overlap.
A second set $F_2$ of CNN filters of $2\times2$ is applied on the resulting $| F_1 | \times n\times 2$ tensor.
Each filter in the second set of filters moves along the embedding dimension one component at a time, as illustrated in \cref{fig:model-annc}, thus resulting in a $|F_2|\times (n-1) \times 1$ tensor. For all CNN filters, the activation function is ReLU.
Finally, the output of the second group of filters is fed to a fully connected layer with a single output that we bind between 1 and 0 using a sigmoid activation.
This final output serves as a classification score, with $0$ for ``not an AP'' and $1$ for valid APs.

Intuitively, the first set of CNN filters extracts the relations $A:B$ and $C:D$ and the second set compares $A:B$ and $C:D$, which can be seen as the predicate ``as'' in ``$A$ is to $B$ as $C$ is to $D$''.
Additionally, while it is possible to assume that embedding dimensions are independent from each other, it is rarely the case in practice.
To handle dependent dimensions, the second set of filters have a size of $2\time2$ which results in overlaps between adjacent embedding dimensions, and the fully connected layer mixes the results of all filters from $F_2$ over all dimensions.

\subsection{Analogy solving using ANNr}\label{sec:model-annr}
\begin{figure}[h]
    \centering
    \includegraphics[width=.95\textwidth]{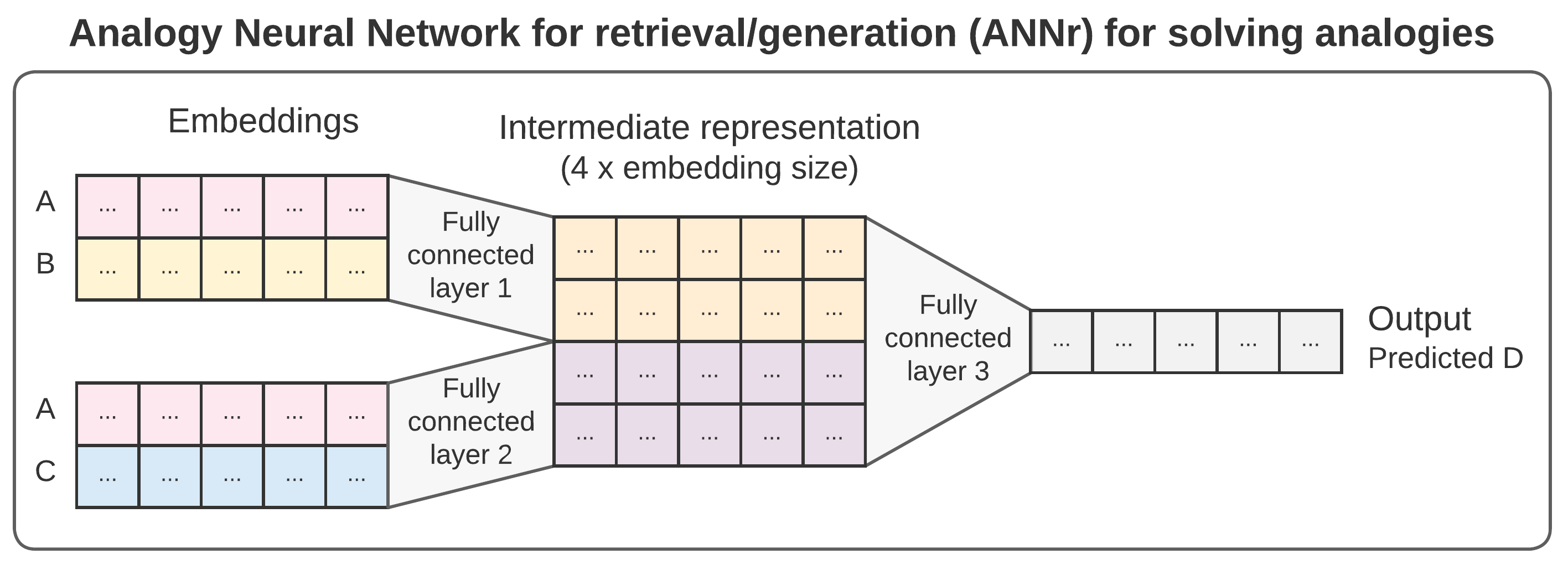}
        \caption{Analogy Neural Network for retrieval/generation (ANNr). The embedding of the three input elements $A,B,C$ are shown horizontally, each in a different color -- notice that $A$ appears twice.}
    \label{fig:model-annr}
\end{figure}
An approach to solving $A:B::C:{x}$ is to find how $e_B$ differs from $e_A$ and to generate an $e_{x}$ that differs from $e_C$ in the same way.
{Central permutation} (see \cref{sec:formal-framework}) allows us to apply the same operations on $A:C::B:{x}$, to obtain $e_{x}$ from $e_B$ using the difference between $e_A$ and $e_C$. The {Analogy Neural Network for retrieval/generation (ANNr)} follows this intuition by using a two step process illustrated in \cref{fig:model-annr}.
First, two separate fully connected layers $f_1,f_2$ with ReLU activation are applied respectively on the concatenation of $A$ and $B$ and the concatenation of $A$ and $C$.
Intuitively, these layer determine the relation between $e_A$ and $e_B$ on the one side while keeping the key content of $e_B$, and similarly for $e_A,e_C$ the other side.
Then, the concatenation of the outputs of $f_1$ and $f_2$ is fed into a last fully connected layer $f_3$ without activation, which generates $e_{x}$ the embedding of the predicted ${x}$.

In preliminary experiments, we observed that using different weights for $f_1,f_2$ achieved better performance than sharing the weights (which would result in a siamese architecture). This indicates a need for asymmetry between the relations between $A$ and $B$ and between $A$ and $C$, and goes in the direction of relaxing the central permutation axiom.

\subsection{Analogy solving with the parallelogram rule}\label{sec:model-parallelogram}
As mentioned in \cref{sec:formal-framework}, the parallelogram is one of the inspirations of the axiomatic setting of APs and has been used to solve APs since early works on word embeddings~\cite{efficient-representation-w2v:2013:mikolov,linguistic-regularities:2013:mikolov}.
To compute the solution of an AP equation $A:B::C:x$, the embeddings $e_A$, $e_B$, and $e_C$ are computed by the embedding model and used to compute $e_x=e_B - e_A + e_C$. Then, $e_X$ is used to find the word $X$ solution to the analogy. With the AE embedding model, $x$ is obtained using the decoder and with the CNN-based model retrieval is used instead.

\subsection{Analogy solving with 3CosAdd or 3CosMul}
3CosAdd~\cite{efficient-representation-w2v:2013:mikolov} and 3CosMul~\cite{3-cos-mul:2014:levy-goldberg} are retrieval approaches to solve AP equations within an embedding space and are often used even if they have known limitations. As retrieval approaches, both methods rely on the embeddings of candidate solutions to solve the equation.
In \cite{analogy-retrival:2022:marquer}, we report results of 3CosAdd and 3CosMul applied on the embeddings trained with ANNc to measure the improvement brought by ANNr.
3CosAdd (\ref{equ:3cosadd}) can be seen as the parallelogram rule combined a cosine similarity to recover the closest solution.
3CosMul (\ref{equ:3cosmul}) maximizes a similarity built upon the cosine. Intuitively, the formula the angle between $B$ and $D$ on one side and the angle between $C$ and $D$ on the other side, and normalizes the result by the angle between $A$ and $D$. In this reading of the formula, 3CosMul shares a similar intuition as ANNr.
To avoid a $0$ denominator, a small $\varepsilon$ is added in (\ref{equ:3cosmul}): 
\begin{align}
    \text{3CosAdd} &= \underset{e_D}{\text{argmax}}\; cos(e_D, e_B-e_A+e_C) \label{equ:3cosadd},\\
    \text{3CosMul} &= \underset{e_D}{\text{argmax}}\; \frac{cos(e_D, e_B)cos(e_D, e_C)}{cos(e_D, e_A) + \varepsilon} 
    \label{equ:3cosmul}.
\end{align}
We refer the reader to~\cite{3-cos-mul:2014:levy-goldberg} for a detailed description of 3CosAdd and 3CosMul.

\subsection{Analogy solving using ANNc}\label{sec:model-annc-ret}
It is also possible to use ANNc to solve analogies, by maximizing the classification score over possible solutions.
This method is most adapted to a retrieval approach, but it is possible to envision generating candidate solutions (for instance, in as was done in Alea) or even to maximize the classification score in a machine learning generation algorithm (for example, using a genetic algorithm).

\section{Pre-training, training and evaluation}
\label{sec:training}
In this section, we first explain in \cref{sec:training-cnn,sec:training-ae} how to pre-train our embedding models, and detail the technical aspects of training our analogy detection and analogy solving models in \cref{sec:training-annc,sec:training-annr} respectively.

Pre-training is a wide-spread approach for embedding models, which consists of two steps: firstly, the embedding model is pre-trained on general-purpose data or data of the target domain, and secondly, the embedding model is transferred to serve as a component of the final model on the target task. For instance, it is possible to transfer a part of a model and reuse it as a component of a larger model, as is usually done with large pretrained embedding models such as Bert~\cite{bert-multilingual:2019:delvin}, wav2vec2~\cite{wav2vec2:2020:develin}, or vision transformers~\cite{vit:2021:dosovitskiy-et-al}.
Then, it is possible to either train only the new part of the final model, but it is also possible to {finetune} the embedding model during the training. 
The pre-training and finetuning approach allows to first bring the embedding model to a globally viable state before specializing it on the task.

Further details on the training hyperparameters are provided in appendix~\cref{app:hparams}.

\subsection{Pre-training the CNN-based embedding model}\label{sec:training-cnn}
In~\cite{analogy-overview-extended-preprint:2021:alsaidi} we confirmed the benefits of pre-training the CNN-based embedding model before analogy solving with ANNr.
The CNN-based embedding is not designed for a standalone pre-training, therefore it is pre-trained on the analogy detection task with an ANNc model, following the ANNc training procedure described below.

\subsection{Pre-training the autoencoder embedding model}\label{sec:training-ae}
This subsection summarizes the task and criterion for pre-training the AE embedding model, following \cite{analogy-genmorpho:2022:chan}.

\paragraph{Autoencoding task.}
AEs are designed to be trained on an autoencoding task (AE task).
The principle of this task is, in our setting, to present a training dataset containing words to the model.
The AE is then trained to encode-decode each word of the training dataset, and to minimize the error in the predicted word.

\paragraph{Training criterion.}
We use the  Cross-Entropy (CE) loss to train the AE, as it is the common way to train a model where each output is one among a series of class.
In the following formula, we write $\mathcal{L}_{AE}$ the loss used for our AE, which is exactly the CE loss but re-written in our setting:
\begin{equation}
\mathcal{L}_{AE} (D, decoder(e_x)) = -\frac{1}{|D|}\sum_{i=1}^{|D|}\sum_{c\in V} D_{c,i} \log{(decoder(e_x)_{c,i})},
\end{equation}
where $|D|$ is the number of characters to predict including the EOS character, $V$ is the character vocabulary (the set of characters present in the training data), $D_{c,i}$ is $1$ it the $i$-th character of $D$ is $c$ and $0$ otherwise,  $decoder(e_x)$ is the decoder output and $decoder(e_x)_{c,i}$ is the predicted {probability} that the $i$-th character of the prediction is $c$.

\subsection{Training ANNc}\label{sec:training-annc}
Here we summarize the training process of ANNc that achieved the most competitive performance, but further details can be found in \cite{detecting:2021:alsaidi,analogy-overview-extended-preprint:2021:alsaidi}.

\paragraph{Data augmentation.} 
The axioms of APs described in \cref{sec:formal-framework} enable the generation multiple valid APs (positive examples) and invalid APs (negative examples) from each AP in our analogy dataset. 
This process is a {data augmentation} process that extends the amount of data available for training, and is necessary to obtain the negative examples required for training the classifier.
Finally, it allows the models to learn how to fit the formal postulates of AP by becoming invariant to them.

By using the \textit{symmetry} and \textit{central permutation} axioms, we are able to generate 7 more valid APs from a valid AP $A:B::C:D$, namely: 
\begin{itemize}
\item $A:C::B:D$, 
\item $D:B::C:A$, 
\item $C:A::D:B$, 
\item $C:D::A:B$, 
\item $B:A::D:C$, 
\item $D:C::B:A$, 
\item $B:D::A:C$.
\end{itemize}
From each of these valid APs, we can generate invalid APs that are in conflict with at least one axiom, following~\cite{analogies-ml-perspective:2019:lim,analogies-ml:2021:lim}.
Given a valid AP $A:B::C:D$, we generate the following invalid APs: $B:A::C:D$, $C:B::A:D$, and $A:A::C:D$, for a total of 24 invalid APs. 
In practice, data augmentation is applied after the embedding model, as shown in \cref{fig:model-summary}.

In the experiments reported in~\cite[Appendix B]{analogy-overview-extended-preprint:2021:alsaidi}, we identified that training ANNc with an imbalance between the positive and the negative examples resulted in a corresponding imbalance in the performance on the two classes of examples.
Additionally, we identified that some invalid APs we generate have the form of valid APs, for instance when using $sang:sang::was:were$ (with the underlying relation of going from the 1st to the 3rd person of singular), the invalid AP generation rule $A:B::C:D \rightarrow A:A::C:D$ produces $sang:sang::was:were$, which we consider valid.
To tackle these issues, we use the following data augmentation process:
\begin{enumerate}
    \item given an AP $A:B::C:D$ we compute the set of 8 valid and the set of 24 invalid APs mentioned above;
    \item we remove APs from the invalid APs that have the same form as any of the valid APs;
    \item if we have more than 8 invalid APs remaining, we sample without replacement 8 invalid APs from the filtered set;
    \item if we have less than 8 invalid APs remaining, we take the filtered set of invalid APs and add the missing amount of APs by randomly sampling them with replacement from the filtered set of invalid APs;
    this process allows us to have at least one instance of each invalid AP.
\end{enumerate}

\paragraph{Training criterion.}
To train ANNc we use the  Binary Cross-Entropy (BCE) loss which is a standard for binary classification, and is written as follows:
\begin{align}
\mathcal{L}_{ANNc} (y, y_{pred}) &= -  y  \log (y_{pred}) - (1 - y) \log (1 - y_{pred})\nonumber\\
&=\begin{cases}
    -  \log (y_{pred}) & \text{ if } y=1,\\
    -  \log (1 - y_{pred}) & \text{ otherwise, }
\end{cases}
\end{align}
where $y$ is the true class ($1$ or $0$) and $y_{pred}$ is the predicted probability for class $1$ (\textit{i.e.}, the probability that the AP is valid, the classification score).

\subsection{Training ANNr}\label{sec:training-annr}
Here we summarize the key elements we use to obtain the best results with ANNr. Further details on the training process can be found in \cite{analogy-retrival:2022:marquer}.

\paragraph{Data augmentation.}
To train ANNr, we use a similar data augmentation process as for ANNc, but considering only the valid APs.
In other words, given an AP equation $A:B::C:x$ with expected $x=D$, we generate 7 other AP equations: \begin{itemize}
    \item 
$A:C::B:x$ with expected $x=D$, 
  \item $D:B::C:x$ with expected  $x=A$,   
  \item $C:A::D:x$ with expected  $x=B$, 
  \item $C:D::A:x$ with expected  $x=B$,
  \item $B:A::D:x$ with expected  $x=C$, 
  \item $D:C::B:x$ with expected  $x=A$, 
  \item $B:D::A:x$ with expected  $x=C$.
\end{itemize}

\paragraph{Training criterion.}
When fine-tuning the embedding model while training ANNr, it is possible to observe collapse of the embedding space if the criterion is badly chosen:
for instance, when using the \textit{mean squared error (MSE)} (or $L^2$ distance, see \cref{equ:mse}), it is possible for the model to move all the embeddings in a smaller area of the embedding space by multiplying them by $10^5$; while this minimizes the MSE, it does not improve retrieval performance as the relative distance between embeddings does not change.

\begin{equation}
    MSE(e_D, e_x) = \frac{1}{n}\sum_{i\in [1,n]}(e_{D,i} - e_{x,i})^2,\label{equ:mse}
\end{equation}
where $e_{D,i}$ is the $i$-th component of $e_D$ and $n$ is the number of dimensions of the embeddings.

In \cite{analogy-retrival:2022:marquer} we experimented with multiple training objectives designed to avoid this phenomena.
In the present work we consider only the most successful one, called $\mathcal{L}_\text{norm. random}$ in \cite{analogy-retrival:2022:marquer} that we rename $\mathcal{L}_\text{ANNr}$ for clarity:
\begin{equation}
\mathcal{L}_\text{ANNr} (e_D, e_x) = \frac{1+MSE(e_D, e_x)}{1 + MSE(\text{batch shuffle}(e_D), e_x)}. \label{equ:random-mse}
\end{equation}
As the model is trained with batches of randomly selected samples, we associate each prediction with an unrelated embedding by permuting $e_D$ along the batch dimension ($\text{batch shuffle}(e_D)$ in \cref{equ:random-mse}).
Intuitively, we encourage the prediction $e_x$ of ANNr to be close to the actual $e_D$ and far from other embeddings by normalizing $\text{MSE}(e_D, {e_x})$ based on distances between other embeddings. 

\paragraph{Inference.}
When used with the CNN-based embedding model, we do not have direct access to the word corresponding to the embedding predicted by ANNr.
We consider all the words appearing in the non-analogical data as candidates for retrieval, which includes words that do not appear in any analogy, and retrieve the closest one based on the cosine distance between the ANNr prediction and the embedding of each candidate word.
Previous experiments~\cite{analogy-overview-extended-preprint:2021:alsaidi} showed that cosine distance slightly outperforms euclidean distance in that setting.

\paragraph{Training ANNr when using the CNN-based embedding model.}
As reported in \cite{analogy-retrival:2022:marquer}, we obtained the best results by first training only ANNr without finetuning the embedding until ANNr converges (\textit{i.e.}, untill there is no improvement in the loss on the devlopement data), then training both models together. We limited the total training time to 50 epochs including both phases.

\paragraph{Training ANNr when using the AE embedding model.}
When training ANNr while finetuning the AE embedding model, we add a generation loss term $\mathcal{L}_{AE}$ to the training loss of the ANNr. In other words, $\mathcal{L}_{AE + ANNr}$ is a convex combination of $\mathcal{L}_{ANNr}$ and $\mathcal{L}_{AE}$.
This allows us to maintain good decoding performance and to make use of the differences between the predicted and expected embedding to guide ANNr at the beginning of training. The $\mathcal{L}_{AE}(e_D, e_x)$ used here is the same as the one used to pre-train the AE:

\begin{equation}
\mathcal{L}_{AE + ANNr} (e_D, e_x) =  (1-\lambda) \mathcal{L}_{ANNr} (e_D, e_x) + \lambda\mathcal{L}_{AE} (D, decoder(e_x)).
\end{equation}

The parameter $\lambda = \min{(\max{(epoch/5, 0.01)}, 0.99)}$ evolves during training, such that at the beginning of the training ($\lambda = 0.99$) ANNr is training almost alone and the AE performance is not altered, and after 5 epochs, only the AE loss is used ($\lambda = 0.01$), following that better performance is obtained by first training only ANNr~\cite{analogy-retrival:2022:marquer}.

In our early experiments, we attempted to use only $\mathcal{L}_{AE} (D, decoder(e_x))$ as the criterion for analogy solving with AE+ANNr, but achieved poor performance. More specifically, the models got stuck in poor local optima at the beginning of training that resulted in the degradation of the quality of the AE while not reaching satisfying analogy solving performance.

\newpage
\part{Empirical study}\label{part:expe}
\label{sec:expe}
In this part we report and discuss the performance of our framework on analogy solving.
We compare multiple variants of our analogy solving approach and two baselines on morphological analogies in 16 languages of the Siganalogies~\cite{siganalogies} dataset we describe in \cref{sec:data}.
This part focuses on latest results on analogy solving presented in \cref{sec:analogy-solving-expe-results}, but we also report additional results on the pre-training of the AE and CNN embedding models in \cref{app:cnn,app:ae}.
For extended studies on the behavior of the ANNc models, we refer the reader to \cite{detecting:2021:alsaidi,transfer:2021:alsaidi,analogy-overview-extended-preprint:2021:alsaidi,transfer:2022:marquer}. We summarize the key findings of these works in \cref{sec:discussion}.

\section{Multilingual morphological analogies data}\label{sec:data}
The Siganalogies~\cite{siganalogies} we proposed in previous work is built upon three datasets described below.
We briefly survey the original datasets as well as the process used to generate analogies.
Siganalogies is a dataset of morphological analogies built upon Sigmorphon 2016~\cite{sigmorphon:2016:cotterell}, Sigmorphon 2019~\cite{sigmorphon:2019:mccarthy-etal}, and the Japanese Bigger Analogy Test Set (JBATS)~\cite{jap-data:2018:karpinska}.
However in our experiments we only consider data from Sigmorphon 2016 and Sigmorphon 2019.
A more detailed description of the Siganalogies dataset, including the number of analogies for each language, can be found on the GitHub page of the datase\footnote{ \url{https://github.com/EMarquer/siganalogies/blob/main/siganalogies_description.pdf}}.

\paragraph{Data from Sigmorphon 2016 and Sigmorphon 2019}
The Sigmorphon shared tasks are a series of shared tasks focusing on tackling morphology-related tasks on multiple languages.

The Sigmorphon 2016~\cite{sigmorphon:2016:cotterell} shared task focuses on \textit{morphological reinflection}, which consists in applying an inflexional morphological transformation on a word that may already be the result of such a transformation.
Taking the example from the authors of the task, going from ``ran'' to ``running'' corresponds to a transformation from (some person of) the past to the present participle of ``to run''.
The dataset covers 10 languages, namely 
Spanish, German, Finnish, Russian, Turkish, Georgian, Navajo, Arabic, Hungarian, and Maltese, which are ``mostly languages with rich inflection'' by the own words of the authors of~\cite{sigmorphon:2016:cotterell}.
The data is extracted from Wikitionary and underwent cleaning and other preprocessing steps.
Sigmorphon 2016 contains 3 tasks, namely \textit{inflection} (task 1), \textit{reinflection} (task 2) and  \textit{unlabled reinflection} (task 3).
In our experiments we focus on the data for the inflection task, which contains triples such as ``run, present participle, running'', although the morphological transformation is specified in more details.
To do so we use a similar process as \cite{sigmorphon:2017:lepage} by combining triples that share the same morphological transformation, for example if we have ``dance, present participle, dancing'' we can combine the two triple to obtain ``run$:$running$::$dance$:$dancing''.
A similar process can be applied to the other two tasks.

The Sigmorphon 2019~\cite{sigmorphon:2019:mccarthy-etal} shared task aims for ``universal morphological inflection'' and covers close to 100 languages.
Once again, 3 tasks are proposed, namely \textit{crosslingual transfer for inflection generation} (task 1) and \textit{morphological analysis and lemmatization in context} (task 2), as well as a task for competing on previous editions of Sigmorphon.
Here we focus only on task 1,  which was to leverage large amounts of data from a high resource language to perform  morphological inflection on a low resource language with little amounts of data.
The data contains pairs of high resource with low resource languages. For each language, the data contains pairs of words and the corresponding morphological transformation, as for task 1 of Sigmorphon 2016.

\paragraph{Data from the Japanese Bigger Analogy Test Set}
The Japanese Bigger Analogy Test Set (JBATS)~\cite{jap-data:2018:karpinska} is an analogy dataset covering 4 categories of relations: derivational and inflexional morphology as well as lexicographic and encyclopedic semantics.
It was initially designed to evaluate the performance of the sub-character and character level embedding models in Japanese.

Both morphological categories contain pairs of words sharing the same morphological transformation, that can be manipulated in a similar way as the Sigmorphon data.
For each second word of the pair, a kanji form and an hiragana/katakana form are provided to cover possible alternate writings.
In previous studies, both writing were considered without distinction, but further experiments in that regard would be interesting in further work.

\paragraph{Siganalogies}
Siganalogies~\cite{siganalogies} covers more than 80 languages, however not all of them are suited for our study.
Our language selection process is done in two steps that we detail in the following paragraphs: we first exclude languages that do not have enough training analogies, then we train an AE model on each of the remaining languages and keep only languages with above 80\% accuracy for the AE task.

We first put aside languages in Sigmorphon 2019 that do not contain enough analogies for training the models, namely the low resource languages as well as Basque and Uzbeck, as they contain less than 50000 analogies in the training set. We also exclude Japanese as it contains kanjis, which require significantly different hyperparameters to achieve acceptable performance.
Note that some languages appear in both Sigmorphon 2016 and Sigmorphon 2019, but are counted as separate languages. Indeed, the pre-processing done in these two editions of Sigmorphon is not the same, in particular with regards to alphabets: in Sigmorphon 2016 everything is transformed in the Latin alphabet while in Sigmorphon 2019 the original alphabet is kept. 

At this step, we have 52 languages remaining: 10 languages from Sigmorphon 2016 and the remaining 42 from Sigmorphon 2019, with 6 languages (Arabic, Finnish, German, Russian, Spanish, and Turkish) appearing in both groups.
The AE is trained on each of the remaining languages, with 5 different training / test split and 10 random initialization per model, for a total of 2600 models ($5 \times 10 \times 50$).
The accuracy of the AE is detailed in \cref{tab:decoder-results} in appendix.

For experiments beyond training the AE on each language, we only consider 10 models trained on the first data split and only languages where the AE achieved an accuracy above 80\% on the AE task. 
The apparent discrepancies between the performance reported in \cref{tab:decoder-results} and the languages we selected (in particular for Portuguese in Sigmorphon 2019 and Russian in Sigmorphon 2016) come from the differences in performance between the first data split and the other 9 splits, which can be appreciated from the large standard deviation reported. 

We end up with 16 languages: Turkish, Hungarian, and Georgian of Sigmorphon 2016, and Adyghe, Arabic, Bashkir, English, French, Hebrew, Portuguese,  Sanskrit, Slovak, Slovene, Swahili, Welsh, and Zulu from Sigmorphon 2019.

\section{Training, development and test data}\label{sec:data-split}
In our experiments, we use the standard of three dataset splits: the training set containing the examples learned by the model, the development set containing examples not seen during training and used to identify overfitting and interrupting the training when the performance stops increasing, and finally the test set to measure the final performance, with examples that appear in neither the training or devlopement set.

For APs, we follow the training and test split from Sigmorphon 2016~\cite{sigmorphon:2016:cotterell}, Sigmorphon 2019~\cite{sigmorphon:2019:mccarthy-etal}, and JBATS~\cite{jap-data:2018:karpinska} when possible, and split the training set into our training and developement analogies.
We exclude duplicates of the form $A:B::C:D \leftrightarrow C:D::A:B$ (corresponding to the axiom of symmetry) but we keep analogies of the form $A:B::A:B$. 
We randomly sample 500 development and 5000 test APs and at most 50000 training APs.

For the AE task, we consider all the words appearing in training and test data from Sigmorphon and JBATS, which includes words that do not appear in any AP (either because the AP is not used, or because there is no two pair of with the same morphological transformation containing the aforementioned word).
We randomly sample 500 development and 500 test examples, and at most 40000 training examples, with no overlaps between the sets.

\section{Analogy solving models}\label{sec:analogy-solving-expe}
Our framework allows us to propose both retrieval-based models and generation-based models, that we compare with two baseline symbolic approaches: the {Alea} approach from~\cite{analogy-alea:2009:langlais} as implemented in~\cite{minimal-complexity:2020:murena}, and the \textit{Kolmo} approach from~\cite{minimal-complexity:2020:murena}.

We consider 3 the retrieval model variants using the same CNN-based embedding model pre-trained using ANNc on analogy detection: 
{CNN+ANNc}, {CNN+3CosMul}, and {CNN+ANNr}.  CNN+ANNc corresponds to re-using both models without fine-tuning on the retrieval task, as mentioned in \cref{sec:model-annc-ret}. CNN+3CosMul uses the 3CosMul~\cite{3-cos-mul:2014:levy-goldberg} approach on the embeddings obtained with the CNN-based model, once again without fine-tuning. Finally, in CNN+ANNr the embedding model is finetuned together with the ANNr, as described in \cref{sec:training-annr}.
The CNN+3CosMul and CNN+ANNc retrieval models serve as a baseline to measure the improvement brought by using ANNr in a retrieval approach.
In \cite{analogy-retrival:2022:marquer} we used the 3CosMul~\cite{3-cos-mul:2014:levy-goldberg} and 3CosAdd~\cite{efficient-representation-w2v:2013:mikolov} approaches as baselines for our CNN+ANNr approach and showed significant improvements when using CNN+ANNr.
However, it can be argued that embeddings trained for analogies using ANNc encode relevant features in a way that is hard to manipulate with 3CosMul and 3CosAdd.
We compare CNN+ANNc with CNN+3CosMul to measure this effect, excluding 3CosAdd which performed worse than 3CosMul in our previous studies.
We also compare CNN+ANNc with CNN+ANNr to ascertain the benefit of training an ANNr model when the setting is limitted to retrieval.

In \cite{analogy-genmorpho:2022:chan} we explored the potential of a character-level AE for analogy solving, using a similar approach as \cite{vec-to-sec-sent-analogies:2020:lepage} that we call {AE+parallel} here.
Following the same approach, we pre-train an AE on our data without analogies then compute the embedding of the solution using the parallelogram rule (\cref{sec:model-parallelogram}), and finally we use the decoder of the AE to generate the predicted solution word.
The last model we experiment with is {AE+ANNr} which combines the AE embedding model with ANNr to achieve higher performance.

\section{Analogy solving performance}\label{sec:analogy-solving-expe-results}
To evaluate the performance of all analogy solving models we use the accuracy.
For retrieval models, this corresponds to top-$1$ retrieval accuracy, additionally hit rate at $k$ for $k\in\{1,3,5,10\}$ are listed in appendix \cref{app:extended-results-retrieval}.
For generative models, we use word accuracy b taking only the most likely prediction for each character, additionally character accuracy and other metrics are available in the detailed model outputs.

The retrieval and generative models performance are reported in \cref{tab:results-retrieval} and \cref{tab:results-generation} respectively, with \cref{tab:results} in appendix comparing the results of both approaches in a single table.
In this work we only consider models trained on a single language, and we discuss the possibility of a multilingual model in \cref{sec:discussion}

\begin{table}[htb]
\begin{center}
\begin{minipage}{\textwidth}
\caption{Analogy solving accuracy (rate of fully well predicted words, in \%) in the retrieval setting (closed world assumption), compared to the best generation model (AE+ANNr).
For models depending on an embedding model, and thus sensitive to random initializations, we report performance as mean $\pm$ standard deviation over 10 random initialization in each setting.
}
\begin{tabular*}{\textwidth}{@{\extracolsep{\fill}}l|ccc|c@{\extracolsep{\fill}}}
\toprule
                                        & \multicolumn{3}{c|}{Retrieval models} & {Best gen. model} \\
Language                                & CNN+ANNr & CNN+3CosMul & CNN+ANNc &  AE+ANNr \\
\midrule
\multicolumn{5}{c}{\textit{Sigmorphon 2016}}\\
Georgian    & \textbf{97.60 $\pm$ 0.23} & 85.58 $\pm$  7.37 &         76.77 $\pm$  9.57  & {87.50$\pm$ 2.08}    \\ 
Hungarian   & \textbf{89.06 $\pm$ 1.71} & 74.89 $\pm$  5.27 &         72.95 $\pm$  8.02  & {90.58$\pm$ 0.63}    \\ 
Turkish     & \textbf{84.75 $\pm$ 2.04} & 52.42 $\pm$  4.33 &         44.43 $\pm$ 13.95  & {79.81$\pm$ 8.58}    \\ 
\midrule
\multicolumn{5}{c}{\textit{Sigmorphon 2019}}\\
Adyghe      & \textbf{93.37 $\pm$ 0.97} & 58.01 $\pm$  8.66 &         80.11 $\pm$  5.10  & {98.50 $\pm$ 0.25}   \\ 
Arabic      & \textbf{72.08 $\pm$ 3.49} & 21.67 $\pm$  5.44 &         40.73 $\pm$  9.93  & {83.99 $\pm$ 1.28}   \\ 
Bashkir     &         57.63 $\pm$ 5.48  & 37.76 $\pm$ 11.12 & \textbf{65.38 $\pm$  6.01} & {95.15 $\pm$ 0.92}   \\ 
English     & \textbf{92.29 $\pm$ 0.91} & 66.83 $\pm$ 15.67 &         65.51 $\pm$  6.25  & {92.29 $\pm$ 4.57}   \\ 
French      & \textbf{93.15 $\pm$ 0.96} & 80.37 $\pm$  7.97 &         62.87 $\pm$ 17.65  & {88.25 $\pm$ 2.24}   \\ 
Hebrew      & \textbf{66.52 $\pm$ 2.59} & 15.47 $\pm$ 10.12 &         36.10 $\pm$  4.28  & {92.50 $\pm$ 0.25}   \\ 
Portuguese  & \textbf{93.12 $\pm$ 1.10} & 58.52 $\pm$ 18.48 &         70.53 $\pm$  9.88  & {93.11 $\pm$ 8.32}   \\ 
Sanskrit    & \textbf{64.18 $\pm$ 2.62} & 33.20 $\pm$  9.34 &         42.59 $\pm$  4.88  & {91.48 $\pm$ 0.78}   \\ 
Slovak      & \textbf{56.23 $\pm$ 4.57} & 49.43 $\pm$  3.13 &         39.58 $\pm$  3.37  & {78.90 $\pm$ 0.86}   \\ 
Slovene     & \textbf{71.99 $\pm$ 2.24} & 57.79 $\pm$  8.59 &         51.88 $\pm$  6.69  & {82.41 $\pm$ 10.98}  \\ 
Swahili     & \textbf{68.56 $\pm$ 6.09} & 44.84 $\pm$  7.77 &         44.46 $\pm$  3.85  & {97.49 $\pm$ 0.21}   \\ 
Welsh       & \textbf{63.80 $\pm$ 3.13} & 47.30 $\pm$  4.67 &         47.58 $\pm$  5.72  & {96.79 $\pm$ 0.23}   \\ 
Zulu        & \textbf{76.59 $\pm$ 2.65} & 58.53 $\pm$  4.42 &         42.56 $\pm$  6.55  & {93.42 $\pm$ 0.73}   \\
\botrule
\end{tabular*}
\label{tab:results-retrieval}
\end{minipage}
\end{center}
\end{table}

\begin{table}[htb]
\begin{center}
\begin{minipage}{\textwidth}
\caption{Analogy solving accuracy (rate of fully well predicted words, in \%) in the generation setting (open world assumption), compared to the best retrieval model (CNN+ANNr).
For models depending on an embedding model, and thus sensitive to random initializations, we report performance as mean $\pm$ standard deviation over 10 random initialization in each setting.
Symbolic baselines Alea~\protect{\cite{analogy-alea:2009:langlais}} and Kolmo~\protect{\cite{minimal-complexity:2020:murena}} were tested in the same setting as our models.
}
\begin{tabular*}{\textwidth}{@{\extracolsep{\fill}}l|c|cccc@{\extracolsep{\fill}}}
\toprule
                                        & Best ret. model & \multicolumn{4}{c}{Generative models} \\
Language                                & CNN+ANNr &  AE+ANNr & AE+parallel & Alea & Kolmo \\
\midrule
\multicolumn{6}{c}{\textit{Sigmorphon 2016}}\\
Georgian    & 97.60 $\pm$ 0.23 & \textbf{87.50$\pm$ 2.08} & \textbf{87.06 $\pm$ 6.28}  & 84.97 & 79.94 \\ 
Hungarian   & 89.06 $\pm$ 1.71 & \textbf{90.58$\pm$ 0.63} & 83.57 $\pm$ 6.63           & 35.24 & 32.07 \\ 
Turkish     & 84.75 $\pm$ 2.04 & \textbf{79.81$\pm$ 8.58} & \textbf{83.03 $\pm$ 14.07} & 42.09 & 39.45 \\ 
\midrule
\multicolumn{6}{c}{\textit{Sigmorphon 2019}}\\
Adyghe      & 93.37 $\pm$ 0.97 & \textbf{98.50 $\pm$ 0.25} & 81.90 $\pm$ 7.35           & 47.94 & 31.25 \\ 
Arabic      & 72.08 $\pm$ 3.49 & \textbf{83.99 $\pm$ 1.28} & 78.40 $\pm$ 12.37          &  2.21 &  3.34 \\ 
Bashkir     & 57.63 $\pm$ 5.48 & \textbf{95.15 $\pm$ 0.92} & 80.38 $\pm$ 18.55          & 22.29 & 29.89 \\ 
English     & 92.29 $\pm$ 0.91 & \textbf{92.29 $\pm$ 4.57} & 65.61 $\pm$ 19.96          & 60.15 & 47.69 \\ 
French      & 93.15 $\pm$ 0.96 & \textbf{88.25 $\pm$ 2.24} & 76.04 $\pm$ 10.20          & 54.48 & 54.39 \\ 
Hebrew      & 66.52 $\pm$ 2.59 & \textbf{92.50 $\pm$ 0.25} & \textbf{91.16 $\pm$ 7.29}  & 19.50 & 16.17 \\ 
Portuguese  & 93.12 $\pm$ 1.10 & \textbf{93.11 $\pm$ 8.32} & 62.88 $\pm$ 24.26          & 78.01 & 71.28 \\ 
Sanskrit    & 64.18 $\pm$ 2.62 & \textbf{91.48 $\pm$ 0.78} & 83.83 $\pm$ 5.37           & 42.80 & 28.83 \\ 
Slovak      & 56.23 $\pm$ 4.57 & \textbf{78.90 $\pm$ 0.86} & 82.62 $\pm$ 6.66           & 30.66 & 28.81 \\ 
Slovene     & 71.99 $\pm$ 2.24 & \textbf{82.41 $\pm$ 10.98} & \textbf{80.95 $\pm$ 8.19} &  2.64 &  5.43 \\ 
Swahili     & 68.56 $\pm$ 6.09 & \textbf{97.49 $\pm$ 0.21} & 81.94 $\pm$ 21.80          & 60.23 & 43.02 \\ 
Welsh       & 63.80 $\pm$ 3.13 & \textbf{96.79 $\pm$ 0.23} & 87.62 $\pm$ 12.69          & 14.47 & 19.15 \\ 
Zulu        & 76.59 $\pm$ 2.65 & \textbf{93.42 $\pm$ 0.73} & 81.96 $\pm$ 15.81          & 26.17 & 27.69 \\ 
\botrule
\end{tabular*}
\label{tab:results-generation}
\end{minipage}
\end{center}
\end{table}

\paragraph{General observations.}
A first striking result is that the embedding based approaches outperform the symbolic baselines in all settings using ANNr and in most settings with CNN+3CosMul, CNN+ANNc, and AE+parallel. 
This result is not surprising, as there is a well known trade-off between the performance of deep learning and the explainability of symbolic approaches.
In particular, the contribution of Kolmo~\cite{minimal-complexity:2020:murena} to explainability is clear as it explicitly provides the underlying transformation used to solve the analogy.
However, the contribution of Alea is more subtle as its main contribution in that regard are the theoretical grounding of the method and the ability to limit the space of solutions.

\paragraph{Performance of the symbolic baselines.}
The baseline generation models have very low performance on Arabic, while the deep learning models do not appear to suffer from the same effect.
As reported in~\cite{analogy-genmorpho:2022:chan}, further analysis of the Arabic data reveals that the character encoding used decomposes each character into multiple encoded characters, resulting in longer and more complex sequences of characters than expected. 
However, on Slovene the performance is also very low even if the characters involved are more or less the same as for Portugese and English.
We propose the hypothesis that the languages where Alea and Kolmo perform poorly contain data with irregularities.
Indeed, the data used to test Kolmo and Alea in \cite{minimal-complexity:2020:murena}, introduced in \cite{sigmorphon:2017:lepage}, has been filtered to exclude analogies that can not be solved without knowledge of the language, in other words, without knowledge of the irregularities in the morphology of the language.

Interestingly, English and Portugese where AE+parallel performs the worst are also the languages of Sigmorphon2019 where the symbolic baselines Alea and Kolmo perform the best. Portugese has a very regular inflexional morphology~\cite{morpho-portugese:1981:brakel}, and the design of Alea and Kolmo allows them to handle morphological transformations which are frequent in those two languages, in particular affixation~\cite{morpho-portugese:1981:brakel,analogy-alea:2009:langlais,minimal-complexity:2020:murena}. Given that the performance of AE+ANNr and of models using the CNN-based embedding does not show the same tendency as AE+parallel, we hypothesize that the AE together with the AE pre-training task have trouble handling those transformations.
However, due to the high variance in the performance it is hard to draw definite conclusions.

While it would be beneficial for the understanding of the limitations of Alea and Kolmo to perform an in depth morphological analysis of the data of languages were the performance is low, this falls out of the scope of this work.

\paragraph{ANNr improves on analogy solving performance}
The AE+parallel, CNN+ANNc, and CNN+3CosMul models have significantly lower performance on analogy solving than CNN+ANNr and AE+ANNr.
While it is possible that this performance gap is caused by the finetuning on analogy solving present when ANNr is used but not the other models, we argue that this is not the main factor.
Indeed, CNN+ANNc and CNN+3CosMul use an embedding model trained with analogy while AE+parallel does not, yet the latter outperforms the former.
By considering the gap in performance between ANNr and the other models, we conclude that ANNr, which was designed and trained specifically for analogy solving, performs better on that task than general purpose embedding models.

As can be seen in \cref{tab:results-generation}, we obtain equivalent or better performance with AE+ANNr than we obtained from CNN+ANNr, even though the latter benefits from an embedding model designed specifically for morphology and from a closed set of possible solutions to retrieve from.
The training procedure and architecture of AE+ANNr compared to CNN+ANNr differ mostly in the use of the decoder output to compute the loss in AE+ANNr. By having direct access to the characters corresponding to the output of ANNr, it is likely that the model learns to avoid small differences in the word form that are not well captured by the CNN-based embedding model. In depth comparison of the errors of the two models would be required to confirm this hypothesis. We leave such studies for further work. 

\paragraph{Analogical training reduces sensitivity to random initialization.}
Putting aside the actual performance achieved, we observe that using ANNr significantly reduces the standard deviation of model performance compared to using only the pretrained AE embedding. Indeed, as shown in appendix \cref{tab:decoder-results}, the AE has a very large standard deviation on the AE task, even when considering 50 models. A similar observation can be made in \cref{tab:results-retrieval,tab:results-generation} for CNN+3CosMul, CNN+ANNc, and AE+parallel, while CNN+ANNr and AE+ANNr have a significantly lower standard deviation.

There are only two differences between the two groups of models: the use of ANNr and the finetuning of the embedding models using analogical data augmentation.
In  \cite{transfer:2021:alsaidi}, we also identified the significant importance of data augmentation in the performance and stability of CNN+ANNc models on morphological analogies.
Moreover, we observed on a different task in (the target sense verification~\cite{analogy-tsv:2022:zervakis}) that using ANNc and analogy during training also reduces the sensitivity of the model to input encoding.
Overall, it appears that analogical training with ANNc or ANNr allows to compensate for high variance in embedding model performance and sensitivity to input conditions.  

\paragraph{Use of ANNc for analogy solving.}
Our experiments with ANNc revealed an unexpected weakness of the approach, namely, the time required to compute the score.
For instance, for Turkish from Sigmorphon2016, in average, running the retrieval with CNN+ANNc took more than 20 minutes against 46 seconds for CNN+3CosMul and 40 seconds for CNN+ANNr. In comparison, it took 2.5 minutes to train for CNN+ANNc and 2 minutes for CNN+ANNr (4.5 minutes if we include the pretraining of the CNN-based embedding model), on a computer equipped with an Nvidia RTX A5000 Mobile used at close to 100\% of its processing capabilities.
In this example, the time required to retrieve the solution with CNN+ANNc is an order of magnitude 10 times longer than the time to train an CNN+ANNr models from the CNN+ANNc model and apply it. Using CNN+3CosMul is even faster as it is not necessary to train a new model, reaching a 30 times speedup.
This issue is caused by having to repeat the computation of the classification score for all the words of the vocabulary for each analogy to solve, however careful engineering might reduce the impact on run time. For example, selecting a subset of candidates using 3CosMul before applying CNN+ANNc as a retrieval model would allow for a significant speedup.

Additionally, the ANNc model is designed for classification and has a tendency to distinguish poorly between comparatively meaningful solution to the AP equation.
This can be confirmed by extending the results to the top 10 highest classification scores, which significantly increases the accuracy as shown in appendix \cref{app:extended-results-retrieval}.

Despite these limitations, CNN+ANNc outperforms CNN+ANNr on Bashkir, and CNN+3CosMul on 7 languages (Adyghe, Arabic, Bashkir, Hebrew, Portugese, Sanskrit, and Welsh).
As mentioned before, the results become even more interesting when we look at the performance within the top 3, 5, and 10 retrieved words, for which CNN+ANNc outperforms CNN+3CosMul respectively on 8, 11, and 12 languages out of 16. This result is in line with the properties of ANNc and 3CosMul for retrieval, as 3CosMul relies on an arbitrary retrieval formula, while ANNc is learned together with the embedding, the latter resulting in a more appropriate alignment between the embedding space and the retrieval approach used.
In simpler terms, ANNc is what was used to train the CNN-based embedding model for morphological analogies, so the features learned by the CNN-based embedding model are easier to properly manipulate with ANNc than with another model.

\section{Discussion}\label{sec:discussion}
In this section we discuss the results of previous sections as well as the ones of our previous papers on the proposed approach, and also provide recommendations for using and further developing the framework.

\subsection{Benefit of ANNc and ANNr over symbolic, non-parametric, and other parametric approaches}
The framework we propose for analogy detection and solving uses ANNc and ANNr, two deep learning models inspired from the properties of APs.
It can be argued that the architecture of ANNc and ANNr is not the most fitting for the manipulation of APs, yet it is relatively easy to grasp and achieves good performance.
As shown in this paper, the performance achieved is better than the one of non-parametric models such as the parallelogram rule and 3CosMul. Additionally, \cite{analogies-ml:2021:lim} showed, for semantic APs, the benefit of the ANNc and ANNr models over multi-layer perceptron~\cite{neural:1994:haykin} (up to 5 layers), random forest~\cite{random-forest:1995:ho} and support vector machine~\cite{svm:1995:cortes}.
The performance improvement is especially striking for ANNr, while for ANNc the benefits are seen mostly on the SAT-based task which is considered a harder semantic analogy task~\cite{analogies-ml:2021:lim}.

Our framework also outperforms symbolic approaches to morphological APs.
In all our works we compare ourselves with Alea~\cite{analogy-alea:2009:langlais} and Kolmo~\cite{minimal-complexity:2020:murena}, as well as with the approach of~\cite{tools-analogy:2018:fam-lepage} in some cases.
In \cite{analogy-overview-extended-preprint:2021:alsaidi} we compare CNN+ANNc with~\cite{tools-analogy:2018:fam-lepage} as well as a relaxed formulation of Alea and Kolmo. It can be argued that comparing generative approaches to analogy solving with analogy detection approaches such as Alea and Kolmo is unfair, and this was taken into account before drawing conclusions. Additionally, the top 10 generated solutions were considered in particular for valid AP detection, resulting in a higher accuracy for the baselines. In this setting, ANNc very significantly outperforms the baselines on valid APs with no significant difference on invalid APs. 

AE+ANNr and CNN+ANNr outperform the symbolic baselines by a very significant margin in all cases.
Furthermore, by considering the 10 most likely retrieval results for CNN+ANNr (see \cref{app:extended-results-retrieval}), the accuracy improves above 99\% for all but 6 languages above 95\%. We observed a similar phenomenon in~\cite{analogy-overview-extended-preprint:2021:alsaidi} for CNN+ANNr, with above 98\% accuracy for all but 2 languages (at 95\% and 92\% respectively) when considering the 10 most likely results.

In~\cite{analogy-genmorpho:2022:chan} we managed to significantly outperform our symbolic baselines with AE+parallel and to reach performance slightly lower performance with the generation approach than with the retrieval approach CNN+ANNr. Further analysis of the results revealed that AE+parallel was highly accurate with regular morphology and with certain permutations, while it struggled when the morphology was more irregular. These considerations resulted in us developing AE+ANNr, which outperforms CNN+ANNr on multiple languages as shown in \cref{tab:results-generation}.

\subsection{Improvements in the training process}
To achieve the best possible performance, we experimented with many different parameters of the training procedure to reach the current state of the framework.

Firstly, regarding ANNc, in \cite{detecting:2021:alsaidi} we identified an imbalance in the analogy detection performance due to the data augmentation process, as we later identified in \cite{analogy-overview-extended-preprint:2021:alsaidi}: using the 8 equivalent forms of a valid analogy together with the 24 corresponding invalid forms skewed the data in favor of invalid analogies, while using only the 3 invalid forms obtained before computing the 8 valid form skewed the data in the other direction, resulting in a matching imbalance in the performance over the two classes for CNN+ANNc. A compromise was found by sampling 8 out of the 24 invalid forms, as presented in \cref{sec:training-annc}.
Based on detailed analyses of results of some languages, in particular from \cite{transfer:2022:marquer,analogy-retrival:2022:marquer}, we further refined the data augmentation to exclude invalid forms that also appear in the valid forms.
As explained later and as can be seen in \cite{transfer:2022:marquer,analogy-tsv:2022:zervakis,medical-analogies:2022:alsaidi}, proper care should be given to the data augmentation process to make the most of the benefits it brings.

Secondly, the current training procedure of ANNr is the result of multiple observations from \cite{analogy-retrival:2022:marquer}. Indeed, pre-training the embedding is necessary to achieve good performance on analogy solving with ANNr, and this can be achieved with relative ease by using CNN+ANNc (with a 10\% to 40\% reported in~\cite{analogy-retrival:2022:marquer}) or an AE task (as we identified in preliminary experiments to the present work). We also observed that when training only ANNr, the model performance was not satisfying, yet naively training CNN+ANNr with MSE resulted in a collapse of the embedding space. After experimenting with various training criterion based on these results, we achieved the best results with $\mathcal{L}_{ANNr}$ (as currently presented in \cref{sec:training-annr}).
While $\mathcal{L}_{ANNr}$ is an easy to implement way to mitigate embedding space collapse, in many settings it achieves comparable performance to the other training criterion we considered, so we encourage the reader to consider \cite[Table 1.]{analogy-retrival:2022:marquer} to have a better outlook when choosing the training criterion.

Other details, in particular with regards to the implementation of approach, were refined along our experiments.
It would be beneficial to further refine our training process as well as the training parameters and hyperparameters, but we leave those considerations for further work due to time limitations.

\subsection{Multilinguality}
In this work we did not attempt to model multiple languages at once or to transfer models between languages, as these aspects were extensively studied for CNN+ANNc in \cite{detecting:2021:alsaidi,transfer:2021:alsaidi,transfer:2022:marquer}.
In particular, in \cite{transfer:2021:alsaidi} we experimented with different forms of transfer from a source language to a target language without finetuning, within Sigmorphon 2016 and JBATS.
Transferring the whole CNN+ANNc resulted in reasonable performance (often above 50\% accuracy), however the transfer failed on some languages. Indeed, those languages used a different set of characters than the rest of languages, resulting in a significant amount of unrecognized characters. To tackle this issue that we call the \textit{alphabet gap}, we used the CNN-based embedding model of the target language together with the ANNc of the source language. This did not improve the performance for all languages, likely due to a mismatch between the embedding space of the  CNN-based embedding model (trained on the target language) and the one of the ANNc (trained on the source language).

In  \cite{transfer:2021:alsaidi} we also experimented with a model covering multiple languages: either two representative languages or all languages at once, with several variants. 
All variants achieved comparable results, with a high performance on positive examples and a poorer performance on negative examples. In all cases, as expected, the overall performance was poorer than with models dealing with only one language, with a slightly higher performance for bilingual models and a more consistent performance for the models on all languages.

We performed further transfer experiments in \cite{transfer:2022:marquer} by transferring the full CNN+ANNc on Sigmorphon 2019, which covers a wider range of languages.
We leveraged languages present in both Sigmorphon 2016 and Sigmorphon 2019 to confirm that our approach generalizes even when the distribution of morphological transformations is not the same, with a strong correlation between the performance loss of transferred models and the alphabet gap.
A striking example is that of Arabic, for which the character representation in Sigmorphon 2016 is completely different than that of Sigmorphon 2019: the former uses a romanized version of the words while the latter uses the actual UTF8 encoding of the Arabic characters.
Furthermore, the high-resource languages of Sigmorphon 2019 form 4 clusters (plus 4 outliers) corresponding to different sets of characters, \textit{i.e.}, different alphabets, as identified in~\cite{transfer:2022:marquer}.
Within each cluster and in particular the largest one, some alphabet gap remains (\textit{e.g.}, the accents present in French but not English), however it is no longer correlated with the transfer performance.
Instead, we found striking similarities between the latter and the language families that can be found in Wikipedia.  In particular, similar clusters were found in a hierarchical clustering based on the transfer performance and the hierarchy of language families.

Based on our observations, it is possible to use the proximity in the Wikipedia language families to predict the performance of transferred models, which not only confirm the transferability of morphological analogies between languages but also opens possibilities to empirically study morphological similarities between languages.
Moreover, these results can be used to identify the language to use when dealing with low resource languages, by finding a closely related high resource language to train the language on.
Finally, the large scale results with comparable transfer performance between closely related languages can be used to design a multilingual analogy model similar to \cite{transfer:2021:alsaidi} on a larger scale. One such approach may be studied in further work, for example using different components of the multilingual model for each cluster of languages.

\subsection{Behavior of the models with regards to the axiomatic setting}
In \cite{transfer:2022:marquer}, transfer is also used to confirm that changing the axioms used in the data augmentation process results in a sigificantly different model corresponding to the changed axiomatic settings. 
This highlight the importance of the choice of the axiomatic setting in our framework, in particular since some axioms are discussed in some application settings \cite{medical-analogies:2022:alsaidi,ap:2022:antic}, as the model behavior can significantly change depending on the chosen axioms.
If enough data is available, we hypothesize that it is possible to determine the most fitting set of axioms by finding the axiomatic setting with the closest performance to the analogical data.
Experiments on a wider range of application domains would be required to confirm this hypothesis.

When analyzing the results of Navajo and Georgian in \cite{analogy-retrival:2022:marquer}, we identified several interesting behaviors of CNN+ANNr:
\textit{(i)} there is no significant performance difference based on permutation, which indicates that the model is invariant to the axioms of APs, \textit{(ii)} reflexivity and identity ($A:B::A:B$ and $A:A::B:B$ respectively) are well handled, \textit{(iii)} when an example appears to violate strong reflexivity or strong inner reflexivity (examples like $A=B,C\neq D$ and $A\neq B,C=D$) the ANNr model frequently gives a (wrong) answer which corresponds to a proper application of the violated axiom. 

\subsection{Model input}
In \cite{analogy-tsv:2022:zervakis} we applied the data augmentation process and ANNc to target sense verification, and observed that using ANNc together with our data augmentation during training reduces the sensitivity of the model to input encoding. The results of the present paper also indicate that data augmentation reduces the sensitivity to the initial setting of the model. Overall, it appears that our framework results in models that are less sensitive to slight changes in their input, as they are made invariant to changes in the input due to the axioms of APs.

Additionally, we explored in \cite{analogy-overview-extended-preprint:2021:alsaidi} the sensitivity of CNN+ANNc and CNN+ANNr to perturbations in their input, by applying dropout on the word embeddings.
By randomly replacing some embedding dimensions by 0 with a given probability, we found that beyond a certain dropout probability the performance of CNN+ANNc dropped drasticaly for invalid APs, and observed a similar but less striking phenomenon for CNN+ANNc on valid APs and for CNN+ANNr.
The probability threshold depends on the language we consider, and we observe a plateau of high performance before the threshold. This allowed us to identify languages for which the embedding appears to contain redundant information (hence there is little effect of removing a few) or conversely languages that might benefit from an increase in embedding size (where the plateau of high performance does not appear, and that are more sensitive to dropout). In the latter category we find Japanese, which is in line with the larger set of possible characters.

\subsection{Limitations of the data used}
Over our experiments, we identified several limitations of our work due to the data we use for training our models.

Firstly, the encoding of characters used in Sigmorphon 2016 is significantly different from the one of Sigmorphon 2019 and JBATS, as they respectively a rewriting of the language using roman alphabet and the standard UTF8 encoding of the characters of the language. As discussed above, this difference results in an alphabet gap between some languages, limiting the possibility of transferring models
Additionally, JBATS contains multiple writings for some words, which has not been leveraged in our previous work on that language.

Secondly, for most languages in the Siganalogies dataset, the grammatical categories represented are very skewed towards specific types of words.
Among other effects, this impacts the generalization ability of the AE embedding model as can be seen when trying to apply the model on words of grammatical categories underrepresented in the data.
On the flip-side, from the results of \cite{transfer:2022:marquer} it appears that CNN+ANNc overcomes this limitation and is able to properly generalize, if we exclude the impact of the alphabet gap.

It is interesting to note that \cite{minimal-complexity:2020:murena} uses a slightly different version of the Sigmorphon 2016 data, where the analogies that cannot be inferred by relying  only on the characters of the source terms have been removed.

\subsection{Limitations of the symbolic approaches}
Along our experiments, and in particular in \cite{analogy-retrival:2022:marquer}, we identified several limitations of the Alea~\cite{analogy-alea:2009:langlais} and Kolmo~\cite{minimal-complexity:2020:murena} symbolic approaches that we use as baselines in most of our experiments.
For instance, longer words and words with many repeated adjacent letters significantly reduce the speed of the two approaches, with this phenomenon particularly striking for Kolmo, as was already identified by the authors of \cite{minimal-complexity:2020:murena}. 
We solved this issue by introducing a timeout for Alea and Kolmo as described in \cite{analogy-retrival:2022:marquer}, as it appears that for most languages less than 0.5\% of the AP equations took longer than 10 seconds to solve. 
Additionally, in \cite{detecting:2021:alsaidi} we found that Kolmo struggled with APs obtained in the data augmentation by using central permutation. It is interesting that this approach, based on empirical observations of human behavior in analogy solving, does not follow all the axioms of APs, as it further highlights the need for flexibility with regards to the axiomatic setting.

\subsection{Diversity in tackling analogical proportions}
Our framework proposes multiple technical solutions, in particular for analogy solving, each with their own benefits.
For instance, while ANNr outperforms 3CosMul, the latter does not require training an additional model. Also, while using ANNc for retrieval outperforms 3CosMul without requiring additional training, using the former is not recommended in most cases due to two key limitations: \textit{(i)} the execution time for retrieval is greater than that of training an ANNr model and using traditional cosine-based retrieval, and \textit{(ii)} ANNc tends to give similarly high scores to multiple solutions, which entails the expected solution appearing further in the ranking.
However, \textit{(ii)} can be a desired property of the model, and \textit{(i)} can be reduced by engineering the process and preprocessing the candidates more effectively. 
Additionally, as reported in \cite{analogy-retrival:2022:marquer}, 3CosMul does not systematically perform better than 3CosAdd, so it can be relevant to test both models.
Similarly, while we use cosine distance for retrieval with CNN+ANNr based on preliminary experiments, there was no significant difference between cosine and Euclidean distance.

Beyond the choice of the model, using our framework requires to formulate the problem to tackle as APs.
This step can be challenging, as can be seen in \cite{medical-analogies:2022:alsaidi} or \cite{analogy-tsv:2022:zervakis} where multiple formulations are explored, but the effort is usually rewarded by good performance and the integration of analogical knowledge in the model.

\section{Conclusion}
\label{sec:ccl}
Our framework achieves very high analogy detection and solving performance, in a variety of settings.
Both the CNN+ANNc for analogy detection and the AE+ANNr for analogy solving achieve consistent state of the art performance on all the languages used in our experiments, and outperform the symbolic baselines Alea~\cite{analogy-alea:2009:langlais} and Kolmo~\cite{minimal-complexity:2020:murena} by as much as 80\% in some cases.
This gain in performance results from, among other things, \textit{(i)} a representation of words learned on the data for the purpose of analogy manipulation, \textit{(ii)} the ability of the model to integrate the dependencies between the embedding dimensions, and \textit{(iii)} the flexibility to go beyond arbitrary arithmetic formulas of APs such as 3CosAdd, 3CosMul, or the parallelogram rule.
However, there is a known trade-off between the performance of deep learning models and their interpretability. In particular, it is usually difficult to understand why an artificial neural network obtains a particular output. This also applies to our framework, but further work might provide theoretical guaranties or empirical methods to tackle this limitation.

In the current work we performed two sets of new experiments to confirm the benefit of using ANNc and ANNr.
Firstly, we compared the retrieval performance of ANNc to ANNr and 3CosMul and showed that ANNc can be used to outperform 3CosMul, in particular when considering the top-10 retrieved solutions.
However, using ANNc in this manner results in significant increase in retrieval time, which should be addressed in further work.
Secondly, we combined ANNr and the AE embedding model, and outperform the symbolic baselines but also the previous approach relying on the parallelogram rule by as much as 30\% in some cases.
This new model outperforms even the retrieval approach for multiple languages, despite the latter benefiting from a closed set of possible solutions.

Overall, our framework shows that it is possible to obtain high performance when manipulating APs beyond arithmetic models on embeddings or manually-designed approaches.
The framework leverages the properties and intuitions of APs in the design of the models, but also to augment data in a way that benefits model performance, sensitivity to initial conditions, and makes the model invariant with regards to permutations following the axioms of APs.
The approach can be applied to other types of data following our guidelines, as can be seen in \cite{analogies-ml-perspective:2019:lim,analogies-ml:2021:lim,analogy-tsv:2022:zervakis,medical-analogies:2022:alsaidi}.

%% file: appendix.tex
\newpage
\section{Additional experimental results}
This appendix contains additional details for some of our experiments.
\Cref{app:hparams} covers the key hyperparameters used when training the model.
\Cref{app:sem-emb} reports the results of pre-trained semantic embedding on solving morphological AP equations on English.
\Cref{app:cnn} detains the performance of the CNN+ANNc model on analogy detection (the pre-training task of CNN+ANNr and CNN+3CosMul), and \cref{app:ae} covers the pre-training performance of the AE embedding model.
Finally, \cref{app:extended-results-retrieval} provide extended results for the analogy solving models.

\subsection{Training hyperparameters}\label{app:hparams}
The optimisation algorithm used for training our models is as follows:
\begin{itemize}
    \item CNN+ANNc: Adam~\cite{adam:2014:kingma} with a learning rate of $10^{-3}$ and batches of 256;
    \item CNN+ANNr: Adam with a learning rate of $10^{-3}$ and batches of 256;
    \item AE pretraining: NAdam~\cite{nadam:2016:dozat} (slightly better results than Adam in our preliminary experiments) with a learning rate of $10^{-2}$ and batches of 2048;
    \item AE+ANNr: Adam with a learning rate of $10^{-3}$ and batches of 512.
\end{itemize}

\subsection{Semantic pre-trained embedding models on morphology}\label{app:sem-emb}
Here we report the performance of ANNc using pre-trained semantic embeddings, in particular GloVe6B \footnote{GloVe trained on a wikipedia dump, see \url{https://nlp.stanford.edu/projects/glove/}} and fasttext \footnote{\url{https://dl.fbaipublicfiles.com/fasttext/vectors-wiki/wiki.en.vec}}. For GloVe6B we use the models with 50, 100, 200, and 300 dimensions, and the fasttext embeddings have 300 dimensions.
It would be unreasonable to expect the best performance from semantic embeddings on a morphological task, due to for instance synonyms for which the forms (and thus the morphemes) are not distinguished.
However, in English as in many languages some semantic features translate directly to morphemes, for instance the plural ``-s'', which may result in an amount of success.

A first issue encountered is that the pre-trained models cover only a small part of the word vocabulary: 37.69\% for GloVe6B and 53.35\% for fasttext.
Taking only the words covered by both models (to get a comparable result) reduces the vocabulary to 37.15\% of its original size.
Analogies containing only covered represent 10.33\% of the original APs. 
To obtain results that are more comparable to our CNN+ANNc model, which has the embedding trained for AP detection, we  add a fully connected layer with output dimension 80 after the pre-trained embedding model, resulting on embeddings of comparable dimension to the CNN-based ones. Additionaly, this layer is meant to handle finetuning for the embedding model.
We present in \cref{tab:sem-emb-full} results for model trained and tested on the original data, and in \cref{tab:sem-emb-covered} for models using only covered APs.
In each case we specify the dimension of the Glove6B model used, and models with a ``+'' sign are equipped with the fully connected layer. 

As one could expect, models using pre-trained embeddings and trained and tested on non-covered analogies have a significantly lower performance than CNN+ANNc, as shown in \cref{tab:sem-emb-full}, for the most part due to unrecognized words during training and testing. In this setting, there is no significant difference between the models, with improvements of 2\% to 4\% on the F1 when learning a fully connected layer after the embedding.
If we  consider only covered APs for training and testing the model, the F1 for GloVe6B jumps to 94\% to 99\% for models with the fully connected layer and to 81\% to 84\% for models without.
Surprisingly, fasttext, which is based on embeddings of sub-words (which are similar to morphemes), achieves a lower F1 of 52\% without and 56\% with the fully connected layer.

In addition to the advantages of character-based embeddings (avoiding unknown words) and to the performance gap, our morphological embedding approaches result in more lightweight embedding models, since it is not necessary to store an embedding for each word.
For instance, our CNN-based embedding model for English weights less than 100 KB, while Glove6B weights 171 MB, 347 MB, 693 MB and 1 GB for 50, 100, 200, and 300 dimensions, respectively, and fasttext weights 6.6 GB.


\begin{table}[h]
\centering
\caption{Performance of the ANNc model on analogy detection (in \%, mean $\pm$ std.) on English (Sigmorphon 2019), for all analogies.
We consider several {pre-trained} embedding variants, for 5 random initialization of the model.
Models with a ``+'' sign have a fully connected layer after the embedding and before ANNc.
The balanced accuracy (Bal. acc.) is the average of the True Positive Rate (TPR) and True Negative Rate (TNR), weighted by the number of the positive and negative examples respectively.}\label{tab:sem-emb-full}
\begin{tabular}{lcccc}
\toprule
 Model &             F1 &  Bal. acc. &                TPR &                TNR        \\
\midrule
 FastText &  52.21 $\pm$  0.34 &  66.99 $\pm$  0.28 &  59.69 $\pm$  1.34 &  74.30 $\pm$  1.44 \\
 GloVe6B 50 &  52.61 $\pm$  0.21 &  67.30 $\pm$  0.19 &  60.43 $\pm$  2.33 &  74.17 $\pm$  2.55 \\
 GloVe6B 100 &  52.82 $\pm$  0.21 &  67.48 $\pm$  0.18 &  60.05 $\pm$  1.66 &  74.90 $\pm$  1.78 \\
 GloVe6B 200 &  52.88 $\pm$  0.33 &  67.52 $\pm$  0.25 &  60.57 $\pm$  1.96 &  74.47 $\pm$  1.92 \\
 GloVe6B 300 &  52.70 $\pm$  0.24 &  67.38 $\pm$  0.21 &  60.07 $\pm$  1.57 &  74.68 $\pm$  1.86 \\
\midrule
 FastText+ &  56.22 $\pm$  0.35 &  70.08 $\pm$  0.26 &  61.58 $\pm$  1.75 &  78.58 $\pm$  1.70 \\
 GloVe6B 50+ &  54.38 $\pm$  0.34 &  68.70 $\pm$  0.26 &  59.82 $\pm$  1.37 &  77.57 $\pm$  1.47 \\
 GloVe6B 100+ &  55.20 $\pm$  0.25 &  69.31 $\pm$  0.19 &  60.89 $\pm$  1.10 &  77.74 $\pm$  0.89 \\
 GloVe6B 200+ &  55.54 $\pm$  0.34 &  69.59 $\pm$  0.24 &  62.46 $\pm$  1.53 &  76.72 $\pm$  1.78 \\
 GloVe6B 300+ &  55.96 $\pm$  0.24 &  69.92 $\pm$  0.18 &  63.03 $\pm$  1.71 &  76.81 $\pm$  1.66 \\
\midrule
 CNN+ANNc     &  99.39 $\pm$  0.04 &  99.70 $\pm$  0.05 &  99.76 $\pm$  0.12 &  99.64 $\pm$  0.03 \\
\bottomrule
\end{tabular}
\end{table}

\begin{table}[h]
\centering
\caption{Performance of the ANNc model on analogy detection (in \%, mean $\pm$ std.) on English (Sigmorphon 2019), for covered analogies only.
We consider various variants of pre-trained embeddings, for 5 random initialization of the model, with all models trained and tested with covered analogies only.
Models with a ``+'' sign have a fully connected layer after the embedding and before ANNc.
The balanced accuracy (Bal. acc.) is the average of the True Positive Rate (TPR) and True Negative Rate (TNR), weighted by the number of the positive and negative examples respectively.}\label{tab:sem-emb-covered}
\begin{tabular}{lcccc}
\toprule
 Model &             F1 &  Bal. acc. &                TPR &                TNR        \\
\midrule
FastText &  52.14 $\pm$  0.19 &  66.93 $\pm$  0.11 &  60.06 $\pm$  2.14 &  73.79 $\pm$  2.03 \\
 GloVe6B 50 &  81.70 $\pm$  1.43 &  89.81 $\pm$  0.80 &  95.30 $\pm$  0.90 &  84.32 $\pm$  1.88 \\
 GloVe6B 100 &  84.56 $\pm$  1.27 &  91.26 $\pm$  0.69 &  94.51 $\pm$  0.55 &  88.02 $\pm$  1.42 \\
 GloVe6B 200 &  84.03 $\pm$  0.46 &  90.95 $\pm$  0.28 &  94.36 $\pm$  0.55 &  87.54 $\pm$  0.62 \\
 GloVe6B 300 &  82.50 $\pm$  0.42 &  90.09 $\pm$  0.47 &  94.35 $\pm$  2.52 &  85.83 $\pm$  1.73 \\
\midrule
FastText+ &  56.24 $\pm$  0.30 &  70.08 $\pm$  0.24 &  61.08 $\pm$  1.62 &  79.09 $\pm$  1.44 \\
 GloVe6B 50+ &  94.83 $\pm$  0.86 &  97.15 $\pm$  0.47 &  97.77 $\pm$  0.52 &  96.52 $\pm$  0.68 \\
 GloVe6B 100+ &  98.37 $\pm$  0.17 &  99.06 $\pm$  0.11 &  99.11 $\pm$  0.30 &  99.01 $\pm$  0.19 \\
 GloVe6B 200+ &  98.76 $\pm$  0.15 &  99.34 $\pm$  0.10 &  99.51 $\pm$  0.18 &  99.17 $\pm$  0.12 \\
 GloVe6B 300+ &  98.83 $\pm$  0.15 &  99.36 $\pm$  0.09 &  99.47 $\pm$  0.19 &  99.25 $\pm$  0.14 \\

\bottomrule
\end{tabular}
\end{table}

\subsection{CNN embedding pre-training performance}\label{app:cnn}
We report in \cref{tab:cnn-annc-clf} the performance of the CNN+ANNc model on analogy detection.
The models presented here are the same as the ones used for the CNN+ANNc model for analogy solving, and the CNN-based embedding model was reused for CNN+3CosAdd and finetuned for CNN+ANNr.

\begin{table}[h]
\centering
\caption{Performance of the CNN+ANNc model on analogy detection (in \%, mean $\pm$ std.) for 10 random initialization of the model. The balanced accuracy (Bal. acc.) is the average of the True Positive Rate (TPR) and True Negative Rate (TNR), weighted by the number of the positive and negative examples respectively.}\label{tab:cnn-annc-clf}
\begin{tabular}{lcccc}
\toprule
 Language &             F1 &  Bal. acc. &                TPR &                TNR        \\
\midrule
\multicolumn{5}{c}{\textit{Sigmorphon 2016}}\\
 Georgian   &  98.26 $\pm$  0.09 &  99.18 $\pm$  0.07 &  99.39 $\pm$  0.19 &  98.98 $\pm$  0.09 \\
 Hungarian  &  99.82 $\pm$  0.05 &  99.87 $\pm$  0.04 &  99.80 $\pm$  0.09 &  99.95 $\pm$  0.03 \\
 Turkish    &  99.75 $\pm$  0.08 &  99.83 $\pm$  0.05 &  99.76 $\pm$  0.09 &  99.91 $\pm$  0.04 \\
\midrule
\multicolumn{5}{c}{\textit{Sigmorphon 2019}}\\
 Adyghe     &  97.37 $\pm$  0.05 &  98.94 $\pm$  0.04 &  99.55 $\pm$  0.16 &  98.33 $\pm$  0.08 \\
 Arabic     &  99.96 $\pm$  0.02 &  99.98 $\pm$  0.02 &  99.97 $\pm$  0.04 &  99.98 $\pm$  0.01 \\
 Bashkir    &  93.04 $\pm$  0.07 &  97.07 $\pm$  0.07 &  99.07 $\pm$  0.17 &  95.06 $\pm$  0.07 \\
 English    &  99.39 $\pm$  0.04 &  99.70 $\pm$  0.05 &  99.76 $\pm$  0.12 &  99.64 $\pm$  0.03 \\
 French     &  99.86 $\pm$  0.06 &  99.88 $\pm$  0.06 &  99.78 $\pm$  0.11 &  99.98 $\pm$  0.01 \\
 Hebrew     &  98.33 $\pm$  0.12 &  99.18 $\pm$  0.09 &  99.28 $\pm$  0.21 &  99.09 $\pm$  0.10 \\
 Portuguese &  99.83 $\pm$  0.09 &  99.88 $\pm$  0.09 &  99.82 $\pm$  0.17 &  99.95 $\pm$  0.00 \\
 Sanskrit   &  98.08 $\pm$  0.07 &  99.08 $\pm$  0.05 &  99.20 $\pm$  0.12 &  98.96 $\pm$  0.06 \\
 Slovak     &  96.09 $\pm$  0.15 &  98.02 $\pm$  0.14 &  98.25 $\pm$  0.36 &  97.79 $\pm$  0.13 \\
 Slovene    &  99.65 $\pm$  0.03 &  99.82 $\pm$  0.04 &  99.81 $\pm$  0.09 &  99.83 $\pm$  0.03 \\
 Swahili    &  99.25 $\pm$  0.04 &  99.66 $\pm$  0.02 &  99.75 $\pm$  0.07 &  99.57 $\pm$  0.05 \\
 Welsh      &  98.63 $\pm$  0.10 &  99.33 $\pm$  0.11 &  99.37 $\pm$  0.30 &  99.28 $\pm$  0.12 \\
 Zulu       &  99.48 $\pm$  0.04 &  99.78 $\pm$  0.04 &  99.87 $\pm$  0.08 &  99.69 $\pm$  0.03 \\
\bottomrule
\end{tabular}
\end{table}

\subsection{Autoencoder embedding pre-training performance}\label{app:ae}
We report in \cref{tab:decoder-results} the AE accuracy at word level of the AE embedding model.
The results presented cover 5 random data splits and 10 random initialization of the model per split, resulting in 50 models per language.
Only the 10 random initialization of the model were used for the first random data split were used in our other experiments involving AE+ANNr and AE+parallel.

\begin{table}[h]
\centering
\caption{Accuracy (in \%, mean $\pm$ std.) at the word level, of the AE pre-trained for at most 100 epochs on 40,000 random words, for 5 different training / test splits and 10 random initialization of the model in each case. Languages in bold are the ones selected for further experiments.}
\begin{tabular}{lrclr}
\toprule
Language                            & Accuracy (\%)   &  & Language                             & Accuracy (\%)   \\
\midrule
\multicolumn{2}{c}{\textit{Sigmorphon 2016}}                   &  & \multicolumn{2}{c}{\textit{Sigmorphon 2019}}                    \\
Arabic                              & 75.72$\pm$13.98 &  & \textbf{French}     & 76.04$\pm$10.20 \\
Finnish                             & 73.30$\pm$ 9.96 &  & German                               & 64.98$\pm$16.12 \\
\textbf{Georgian}  & 87.06$\pm$ 6.28 &  & Greek                                & 39.44$\pm$21.68 \\
German                              & 69.50$\pm$15.03 &  & \textbf{Hebrew}     & 91.16$\pm$ 7.29 \\
\textbf{Hungarian} & 83.57$\pm$ 6.63 &  & Hindi                                & 58.08$\pm$32.85 \\
Maltese                             & 77.07$\pm$25.07 &  & Hungarian                            & 61.51$\pm$13.81 \\
Navajo                              & 38.36$\pm$21.41 &  & Irish                                & 10.39$\pm$12.59 \\
\textit{Russian}   & 84.70$\pm$ 6.63 &  & Italian                              & 70.92$\pm$13.52 \\
Spanish                             & 78.36$\pm$15.85 &  & Kannada                              & 0.05$\pm$ 0.16  \\
\textbf{Turkish}   & 83.03$\pm$14.07 &  & Kurmanji                             & 76.52$\pm$ 8.50 \\
\multicolumn{2}{c}{\textit{Sigmorphon 2019}}                   &  & Latin                                & 69.90$\pm$13.80 \\
\textbf{Adyghe}    & 81.90$\pm$ 7.35 &  & Latvian                              & 78.29$\pm$ 5.53 \\
Albanian                            & 51.07$\pm$13.56 &  & Persian                              & 59.14$\pm$23.10 \\
\textbf{Arabic}    & 78.40$\pm$12.37 &  & Polish                               & 67.36$\pm$18.58 \\
Armenian                            & 77.94$\pm$ 4.87 &  & \textbf{Portuguese} & 62.88$\pm$24.26 \\
Asturian                            & 73.71$\pm$26.71 &  & Romanian                             & 62.76$\pm$22.14 \\
\textbf{Bashkir}   & 80.38$\pm$18.55 &  & Russian                              & 67.76$\pm$11.51 \\
Belarusian                          & 61.15$\pm$15.30 &  & \textbf{Sanskrit}   & 83.83$\pm$ 5.37 \\
Bengali                             & 47.78$\pm$25.46 &  & \textbf{Slovak}     & 82.62$\pm$ 6.66 \\
Bulgarian                           & 76.07$\pm$ 7.18 &  & \textbf{Slovene}    & 80.95$\pm$ 8.19 \\
Czech                               & 61.08$\pm$19.61 &  & Sorani                               & 77.43$\pm$11.88 \\
Danish                              & 73.88$\pm$11.70 &  & Spanish                              & 67.86$\pm$21.80 \\
Dutch                               & 63.90$\pm$19.41 &  & \textbf{Swahili}    & 81.94$\pm$21.80 \\
\textbf{English}   & 65.61$\pm$19.96 &  & Turkish                              & 68.03$\pm$12.29 \\
Estonian                            & 66.78$\pm$ 9.02 &  & Urdu                                 & 52.02$\pm$28.24 \\
Finnish                             & 34.68$\pm$22.89 &  & \textbf{Welsh}      & 87.62$\pm$12.69 \\
                                    &                 &  & \textbf{Zulu}       & 81.96$\pm$15.81 \\
\bottomrule
\end{tabular}
\label{tab:decoder-results}
\end{table}

\subsection{Extended analogy solving results}
\label{app:extended-results-retrieval}
In this appendix we provide \cref{tab:results} summarizing the results of all our analogy solving approaches for easier comparison.

\begin{sidewaystable}[!h]
\begin{center}
\begin{minipage}{\textwidth}
\caption{Accuracy (rate of fully well predicted words, in \%) on the analogy solving tasks.
For models depending on an embedding model, and thus sensitive to random initializations, we report performance as mean $\pm$ standard deviation over 10 random initialization in each setting.
Symbolic baselines Alea~\protect{\cite{analogy-alea:2009:langlais}} and Kolmo~\protect{\cite{minimal-complexity:2020:murena}} were tested in the same setting as our models.
}
\begin{tabular*}{\textwidth}{@{\extracolsep{\fill}}l|ccc|cccc@{\extracolsep{\fill}}}
\toprule
                                        & \multicolumn{3}{c|}{Retrieval models} & \multicolumn{4}{c}{Generative models} \\
Language                                & CNN+ANNr & CNN+3CosMul & CNN+ANNc &  AE+ANNr & AE+parallel & Alea & Kolmo \\
\midrule
\multicolumn{8}{c}{\textit{Sigmorphon 2016}}\\
Georgian    & \textbf{97.60 $\pm$ 0.23} & 85.58 $\pm$  7.37 &         76.77 $\pm$  9.57  & \textbf{87.50$\pm$ 2.08} & \textbf{87.06 $\pm$ 6.28}  & 84.97 & 79.94 \\ 
Hungarian   & \textbf{89.06 $\pm$ 1.71} & 74.89 $\pm$  5.27 &         72.95 $\pm$  8.02  & \textbf{90.58$\pm$ 0.63} & 83.57 $\pm$ 6.63           & 35.24 & 32.07 \\ 
Turkish     & \textbf{84.75 $\pm$ 2.04} & 52.42 $\pm$  4.33 &         44.43 $\pm$ 13.95  & \textbf{79.81$\pm$ 8.58} & \textbf{83.03 $\pm$ 14.07} & 42.09 & 39.45 \\ 
\midrule
\multicolumn{8}{c}{\textit{Sigmorphon 2019}}\\
Adyghe      & \textbf{93.37 $\pm$ 0.97} & 58.01 $\pm$  8.66 &         80.11 $\pm$  5.10  & \textbf{98.50 $\pm$ 0.25} & 81.90 $\pm$ 7.35           & 47.94 & 31.25 \\ 
Arabic      & \textbf{72.08 $\pm$ 3.49} & 21.67 $\pm$  5.44 &         40.73 $\pm$  9.93  & \textbf{83.99 $\pm$ 1.28} & 78.40 $\pm$ 12.37          &  2.21 &  3.34 \\ 
Bashkir     &         57.63 $\pm$ 5.48  & 37.76 $\pm$ 11.12 & \textbf{65.38 $\pm$  6.01} & \textbf{95.15 $\pm$ 0.92} & 80.38 $\pm$ 18.55          & 22.29 & 29.89 \\ 
English     & \textbf{92.29 $\pm$ 0.91} & 66.83 $\pm$ 15.67 &         65.51 $\pm$  6.25  & \textbf{92.29 $\pm$ 4.57} & 65.61 $\pm$ 19.96          & 60.15 & 47.69 \\ 
French      & \textbf{93.15 $\pm$ 0.96} & 80.37 $\pm$  7.97 &         62.87 $\pm$ 17.65  & \textbf{88.25 $\pm$ 2.24} & 76.04 $\pm$ 10.20          & 54.48 & 54.39 \\ 
Hebrew      & \textbf{66.52 $\pm$ 2.59} & 15.47 $\pm$ 10.12 &         36.10 $\pm$  4.28  & \textbf{92.50 $\pm$ 0.25} & \textbf{91.16 $\pm$ 7.29}  & 19.50 & 16.17 \\ 
Portuguese  & \textbf{93.12 $\pm$ 1.10} & 58.52 $\pm$ 18.48 &         70.53 $\pm$  9.88  & \textbf{93.11 $\pm$ 8.32} & 62.88 $\pm$ 24.26          & 78.01 & 71.28 \\ 
Sanskrit    & \textbf{64.18 $\pm$ 2.62} & 33.20 $\pm$  9.34 &         42.59 $\pm$  4.88  & \textbf{91.48 $\pm$ 0.78} & 83.83 $\pm$ 5.37           & 42.80 & 28.83 \\ 
Slovak      & \textbf{56.23 $\pm$ 4.57} & 49.43 $\pm$  3.13 &         39.58 $\pm$  3.37  & \textbf{78.90 $\pm$ 0.86} & 82.62 $\pm$ 6.66           & 30.66 & 28.81 \\ 
Slovene     & \textbf{71.99 $\pm$ 2.24} & 57.79 $\pm$  8.59 &         51.88 $\pm$  6.69  & \textbf{82.41 $\pm$ 10.98} & \textbf{80.95 $\pm$ 8.19} &  2.64 &  5.43 \\ 
Swahili     & \textbf{68.56 $\pm$ 6.09} & 44.84 $\pm$  7.77 &         44.46 $\pm$  3.85  & \textbf{97.49 $\pm$ 0.21} & 81.94 $\pm$ 21.80          & 60.23 & 43.02 \\ 
Welsh       & \textbf{63.80 $\pm$ 3.13} & 47.30 $\pm$  4.67 &         47.58 $\pm$  5.72  & \textbf{96.79 $\pm$ 0.23} & 87.62 $\pm$ 12.69          & 14.47 & 19.15 \\ 
Zulu        & \textbf{76.59 $\pm$ 2.65} & 58.53 $\pm$  4.42 &         42.56 $\pm$  6.55  & \textbf{93.42 $\pm$ 0.73} & 81.96 $\pm$ 15.81          & 26.17 & 27.69 \\ 
\botrule
\end{tabular*}
\label{tab:results}
\end{minipage}
\end{center}
\end{sidewaystable}

We describe and compare in the hit rate at $k$ for $k\in\{1,3,5,10\}$ for our retrieval models in \cref{tab:results-at1,tab:results-at3,tab:results-at5,tab:results-at10}.
The hit rate corresponds to the rate of analogies for which the expected answer appears in the top $k$ retrieved answers.
For $k=1$, the hit rate is the accuracy.

\begin{table}[!h]
\begin{center}
\begin{minipage}{\textwidth}
\caption{Accuracy, \textit{i.e.},  hit rate at 1 (in \%, mean $\pm$ std.) at the word level for the retrieval models.
**: highest average performance; *: second highest average performance.}
\begin{tabular}{lllll}
\toprule
     &      &              CNN+ANNr &          CNN+3CosMul &              CNN+ANNc \\
Sigmorphon & Language &                       &                      &                       \\
\midrule
2016 & Georgian   &  \textbf{97.60 $\pm$  0.23} **  &          85.58 $\pm$  7.37  *   &          76.77 $\pm$  9.57      \\
     & Hungarian  &  \textbf{89.06 $\pm$  1.71} **  &          74.89 $\pm$  5.27  *   &          72.95 $\pm$  8.02      \\
     & Turkish    &  \textbf{84.75 $\pm$  2.04} **  &          52.42 $\pm$  4.33  *   &          44.43 $\pm$ 13.95      \\
2019 & Adyghe     &  \textbf{93.37 $\pm$  0.97} **  &          58.01 $\pm$  8.66      &          80.11 $\pm$  5.10  *   \\
     & Arabic     &  \textbf{72.08 $\pm$  3.49} **  &          21.67 $\pm$  5.44      &          40.73 $\pm$  9.93  *   \\
     & Bashkir    &          57.63 $\pm$  5.48  *   &          37.76 $\pm$ 11.12      &  \textbf{65.38 $\pm$  6.01} **  \\
     & English    &  \textbf{92.29 $\pm$  0.91} **  &          66.83 $\pm$ 15.67  *   &          65.51 $\pm$  6.25      \\
     & French     &  \textbf{93.15 $\pm$  0.96} **  &          80.37 $\pm$  7.97  *   &          62.87 $\pm$ 17.65      \\
     & Hebrew     &  \textbf{66.52 $\pm$  2.59} **  &          15.47 $\pm$ 10.12      &          36.10 $\pm$  4.28  *   \\
     & Portuguese &  \textbf{93.12 $\pm$  1.10} **  &          58.52 $\pm$ 18.48      &          70.53 $\pm$  9.88  *   \\
     & Sanskrit   &  \textbf{64.18 $\pm$  2.62} **  &          33.20 $\pm$  9.34      &          42.59 $\pm$  4.88  *   \\
     & Slovak     &  \textbf{56.23 $\pm$  4.57} **  &          49.43 $\pm$  3.13  *   &          39.58 $\pm$  3.37      \\
     & Slovene    &  \textbf{71.99 $\pm$  2.24} **  &          57.79 $\pm$  8.59  *   &          51.88 $\pm$  6.69      \\
     & Swahili    &  \textbf{68.56 $\pm$  6.09} **  &          44.84 $\pm$  7.77  *   &          44.46 $\pm$  3.85      \\
     & Welsh      &  \textbf{63.80 $\pm$  3.13} **  &          47.30 $\pm$  4.67      &          47.58 $\pm$  5.72  *   \\
     & Zulu       &  \textbf{76.59 $\pm$  2.65} **  &          58.53 $\pm$  4.42  *   &          42.56 $\pm$  6.55      \\
\bottomrule
\end{tabular}
\label{tab:results-at1}
\end{minipage}
\end{center}
\end{table}

\begin{table}[!h]
\begin{center}
\begin{minipage}{\textwidth}
\caption{Hit rate at 3 (in \%, mean $\pm$ std.) at the word level for the retrieval models.
**: highest average performance; *: second highest average performance.}
\begin{tabular}{lllll}
\toprule
     &      &              CNN+ANNr &          CNN+3CosMul &              CNN+ANNc \\
Sigmorphon & Language &                       &                      &                       \\
\midrule
2016 & Georgian   &  \textbf{99.54 $\pm$  0.23} **  &          93.34 $\pm$  3.82  *   &          92.07 $\pm$  4.92      \\
     & Hungarian  &  \textbf{97.49 $\pm$  0.42} **  &          89.04 $\pm$  3.87  *   &          87.96 $\pm$  5.84      \\
     & Turkish    &  \textbf{98.20 $\pm$  0.48} **  &          68.29 $\pm$  4.04  *   &          63.64 $\pm$ 12.89      \\
2019 & Adyghe     &  \textbf{99.63 $\pm$  0.13} **  &          74.78 $\pm$  7.80      &          95.93 $\pm$  1.84  *   \\
     & Arabic     &  \textbf{91.33 $\pm$  1.94} **  &          38.61 $\pm$  7.44      &          65.41 $\pm$ 11.60  *   \\
     & Bashkir    &          78.50 $\pm$  3.00  *   &          55.72 $\pm$ 12.65      &  \textbf{83.85 $\pm$  3.96} **  \\
     & English    &  \textbf{98.66 $\pm$  0.29} **  &          76.34 $\pm$ 14.94      &          80.50 $\pm$  4.66  *   \\
     & French     &  \textbf{98.33 $\pm$  0.30} **  &          91.60 $\pm$  4.64  *   &          82.94 $\pm$ 13.10      \\
     & Hebrew     &  \textbf{88.06 $\pm$  2.09} **  &          30.20 $\pm$ 15.80      &          59.69 $\pm$  4.93  *   \\
     & Portuguese &  \textbf{99.06 $\pm$  0.21} **  &          79.18 $\pm$ 17.98      &          91.58 $\pm$  6.02  *   \\
     & Sanskrit   &  \textbf{89.95 $\pm$  1.26} **  &          53.90 $\pm$  8.92      &          66.03 $\pm$  4.88  *   \\
     & Slovak     &  \textbf{87.86 $\pm$  1.56} **  &          67.81 $\pm$  4.08  *   &          64.46 $\pm$  4.00      \\
     & Slovene    &  \textbf{92.28 $\pm$  0.72} **  &          78.08 $\pm$  8.26  *   &          75.92 $\pm$  6.94      \\
     & Swahili    &  \textbf{84.72 $\pm$  4.70} **  &          60.57 $\pm$  7.08  *   &          57.34 $\pm$  5.10      \\
     & Welsh      &  \textbf{88.29 $\pm$  1.92} **  &          71.66 $\pm$  5.04      &          76.67 $\pm$  6.09  *   \\
     & Zulu       &  \textbf{89.66 $\pm$  1.66} **  &          77.92 $\pm$  4.10  *   &          64.51 $\pm$  7.17      \\
\bottomrule
\end{tabular}
\label{tab:results-at3}
\end{minipage}
\end{center}
\end{table}

\begin{table}[!h]
\begin{center}
\begin{minipage}{\textwidth}
\caption{Hit rate at 5 (in \%, mean $\pm$ std.) at the word level for the retrieval models.
**: highest average performance; *: second highest average performance.}
\begin{tabular}{lllll}
\toprule
     &      &              CNN+ANNr &          CNN+3CosMul &              CNN+ANNc \\
Sigmorphon & Language &                       &                      &                       \\
\midrule
2016 & Georgian   &  \textbf{99.73 $\pm$  0.14} **  &          94.92 $\pm$  2.63      &          95.14 $\pm$  3.16  *   \\
     & Hungarian  &  \textbf{98.62 $\pm$  0.29} **  &          92.40 $\pm$  3.29  *   &          91.75 $\pm$  4.70      \\
     & Turkish    &  \textbf{99.28 $\pm$  0.20} **  &          73.95 $\pm$  3.43  *   &          70.90 $\pm$ 11.30      \\
2019 & Adyghe     &  \textbf{99.85 $\pm$  0.07} **  &          79.44 $\pm$  7.22      &          98.08 $\pm$  0.97  *   \\
     & Arabic     &  \textbf{95.19 $\pm$  1.14} **  &          47.04 $\pm$  8.23      &          74.75 $\pm$ 11.07  *   \\
     & Bashkir    &          87.12 $\pm$  2.04  *   &          64.30 $\pm$ 11.48      &  \textbf{89.61 $\pm$  2.40} **  \\
     & English    &  \textbf{99.01 $\pm$  0.22} **  &          79.63 $\pm$ 14.15      &          85.14 $\pm$  3.89  *   \\
     & French     &  \textbf{98.86 $\pm$  0.18} **  &          93.91 $\pm$  3.57  *   &          88.61 $\pm$  9.81      \\
     & Hebrew     &  \textbf{93.13 $\pm$  1.43} **  &          40.07 $\pm$ 17.04      &          69.63 $\pm$  4.61  *   \\
     & Portuguese &  \textbf{99.50 $\pm$  0.17} **  &          85.79 $\pm$ 15.20      &          96.26 $\pm$  3.61  *   \\
     & Sanskrit   &  \textbf{95.90 $\pm$  0.48} **  &          63.92 $\pm$  7.26      &          76.12 $\pm$  4.11  *   \\
     & Slovak     &  \textbf{96.50 $\pm$  0.46} **  &          74.26 $\pm$  4.15      &          74.42 $\pm$  3.83  *   \\
     & Slovene    &  \textbf{97.48 $\pm$  0.28} **  &          84.38 $\pm$  7.26      &          84.41 $\pm$  5.89  *   \\
     & Swahili    &  \textbf{90.08 $\pm$  3.70} **  &          66.97 $\pm$  7.03  *   &          63.30 $\pm$  5.61      \\
     & Welsh      &  \textbf{93.88 $\pm$  1.64} **  &          79.10 $\pm$  4.40      &          85.76 $\pm$  4.72  *   \\
     & Zulu       &  \textbf{92.83 $\pm$  1.21} **  &          84.28 $\pm$  3.62  *   &          73.67 $\pm$  6.59      \\
\bottomrule
\end{tabular}
\label{tab:results-at5}
\end{minipage}
\end{center}
\end{table}

\begin{table}[!h]
\begin{center}
\begin{minipage}{\textwidth}
\caption{Hit rate at 10 (in \%, mean $\pm$ std.) at the word level for the retrieval models.
**: highest average performance; *: second highest average performance.}
\begin{tabular}{lllll}
\toprule
     &      &              CNN+ANNr &          CNN+3CosMul &             CNN+ANNc \\
Sigmorphon & Language &                       &                      &                      \\
\midrule
2016 & Georgian   &  \textbf{99.84 $\pm$  0.07} **  &          96.19 $\pm$  1.55      &          97.31 $\pm$  1.68  *   \\
     & Hungarian  &  \textbf{99.21 $\pm$  0.16} **  &          94.97 $\pm$  2.49      &          95.01 $\pm$  3.36  *   \\
     & Turkish    &  \textbf{99.67 $\pm$  0.09} **  &          79.96 $\pm$  2.71  *   &          79.22 $\pm$  8.81      \\
2019 & Adyghe     &  \textbf{99.95 $\pm$  0.03} **  &          84.02 $\pm$  6.46      &          99.30 $\pm$  0.41  *   \\
     & Arabic     &  \textbf{97.82 $\pm$  0.53} **  &          58.14 $\pm$  8.82      &          84.30 $\pm$  9.44  *   \\
     & Bashkir    &  \textbf{98.50 $\pm$  0.50} **  &          74.58 $\pm$  8.39      &          94.27 $\pm$  0.86  *   \\
     & English    &  \textbf{99.31 $\pm$  0.16} **  &          83.26 $\pm$ 13.18      &          89.99 $\pm$  2.95  *   \\
     & French     &  \textbf{99.25 $\pm$  0.13} **  &          95.94 $\pm$  2.50  *   &          93.62 $\pm$  6.07      \\
     & Hebrew     &  \textbf{96.96 $\pm$  0.78} **  &          54.58 $\pm$ 15.98      &          80.83 $\pm$  3.77  *   \\
     & Portuguese &  \textbf{99.74 $\pm$  0.10} **  &          91.98 $\pm$ 10.44      &          98.55 $\pm$  1.50  *   \\
     & Sanskrit   &  \textbf{98.78 $\pm$  0.24} **  &          75.24 $\pm$  4.99      &          86.48 $\pm$  3.03  *   \\
     & Slovak     &  \textbf{98.77 $\pm$  0.31} **  &          80.71 $\pm$  3.73      &          84.65 $\pm$  3.10  *   \\
     & Slovene    &  \textbf{99.16 $\pm$  0.19} **  &          89.77 $\pm$  5.58      &          91.35 $\pm$  4.13  *   \\
     & Swahili    &  \textbf{95.42 $\pm$  2.17} **  &          75.30 $\pm$  6.64  *   &          71.77 $\pm$  5.76      \\
     & Welsh      &  \textbf{97.56 $\pm$  0.84} **  &          85.97 $\pm$  3.66      &          93.26 $\pm$  2.71  *   \\
     & Zulu       &  \textbf{95.96 $\pm$  0.64} **  &          90.70 $\pm$  2.80  *   &          84.30 $\pm$  5.04      \\
\bottomrule
\end{tabular}
\label{tab:results-at10}
\end{minipage}
\end{center}
\end{table}

{\clearpage}

%% file: sn-main.bbl

\begin{thebibliography}{61}
\ifx \bisbn   \undefined \def \bisbn  #1{ISBN #1}\fi
\ifx \binits  \undefined \def \binits#1{#1}\fi
\ifx \bauthor  \undefined \def \bauthor#1{#1}\fi
\ifx \batitle  \undefined \def \batitle#1{#1}\fi
\ifx \bjtitle  \undefined \def \bjtitle#1{#1}\fi
\ifx \bvolume  \undefined \def \bvolume#1{\textbf{#1}}\fi
\ifx \byear  \undefined \def \byear#1{#1}\fi
\ifx \bissue  \undefined \def \bissue#1{#1}\fi
\ifx \bfpage  \undefined \def \bfpage#1{#1}\fi
\ifx \blpage  \undefined \def \blpage #1{#1}\fi
\ifx \burl  \undefined \def \burl#1{\textsf{#1}}\fi
\ifx \doiurl  \undefined \def \doiurl#1{\url{https://doi.org/#1}}\fi
\ifx \betal  \undefined \def \betal{\textit{et al.}}\fi
\ifx \binstitute  \undefined \def \binstitute#1{#1}\fi
\ifx \binstitutionaled  \undefined \def \binstitutionaled#1{#1}\fi
\ifx \bctitle  \undefined \def \bctitle#1{#1}\fi
\ifx \beditor  \undefined \def \beditor#1{#1}\fi
\ifx \bpublisher  \undefined \def \bpublisher#1{#1}\fi
\ifx \bbtitle  \undefined \def \bbtitle#1{#1}\fi
\ifx \bedition  \undefined \def \bedition#1{#1}\fi
\ifx \bseriesno  \undefined \def \bseriesno#1{#1}\fi
\ifx \blocation  \undefined \def \blocation#1{#1}\fi
\ifx \bsertitle  \undefined \def \bsertitle#1{#1}\fi
\ifx \bsnm \undefined \def \bsnm#1{#1}\fi
\ifx \bsuffix \undefined \def \bsuffix#1{#1}\fi
\ifx \bparticle \undefined \def \bparticle#1{#1}\fi
\ifx \barticle \undefined \def \barticle#1{#1}\fi
\bibcommenthead
\ifx \bconfdate \undefined \def \bconfdate #1{#1}\fi
\ifx \botherref \undefined \def \botherref #1{#1}\fi
\ifx \url \undefined \def \url#1{\textsf{#1}}\fi
\ifx \bchapter \undefined \def \bchapter#1{#1}\fi
\ifx \bbook \undefined \def \bbook#1{#1}\fi
\ifx \bcomment \undefined \def \bcomment#1{#1}\fi
\ifx \oauthor \undefined \def \oauthor#1{#1}\fi
\ifx \citeauthoryear \undefined \def \citeauthoryear#1{#1}\fi
\ifx \endbibitem  \undefined \def \endbibitem {}\fi
\ifx \bconflocation  \undefined \def \bconflocation#1{#1}\fi
\ifx \arxivurl  \undefined \def \arxivurl#1{\textsf{#1}}\fi
\csname PreBibitemsHook\endcsname

\bibitem{measure-inteligence:2019:cholet}
\begin{botherref}
\oauthor{\bsnm{Chollet}, \binits{F.}}:
On the measure of intelligence.
CoRR
(2019)
\end{botherref}
\endbibitem

\bibitem{analogy-making-cognition:2001:mitchel}
\begin{bchapter}
\bauthor{\bsnm{Mitchell}, \binits{M.}}:
\bctitle{Analogy making as a complex adaptive system}.
In: \bbtitle{SFISSC},
pp. \bfpage{335}--\blpage{360}
(\byear{2001})
\end{bchapter}
\endbibitem

\bibitem{FahandarH18}
\begin{bchapter}
\bauthor{\bsnm{Fahandar}, \binits{M.A.}},
\bauthor{\bsnm{H{\"u}llermeier}, \binits{E.}}:
\bctitle{Learning to rank based on analogical reasoning}.
In: \bbtitle{AAAI},
pp. \bfpage{2951}--\blpage{2958}
(\byear{2018})
\end{bchapter}
\endbibitem

\bibitem{FahandarH21}
\begin{bchapter}
\bauthor{\bsnm{Fahandar}, \binits{M.A.}},
\bauthor{\bsnm{H{\"u}llermeier}, \binits{E.}}:
\bctitle{Analogical embedding for analogy-based learning to rank}.
In: \bbtitle{IDA}.
\bsertitle{LNCS},
vol. \bseriesno{12695},
pp. \bfpage{76}--\blpage{88}
(\byear{2021})
\end{bchapter}
\endbibitem

\bibitem{HugPRS19}
\begin{barticle}
\bauthor{\bsnm{Hug}, \binits{N.}},
\bauthor{\bsnm{Prade}, \binits{H.}},
\bauthor{\bsnm{Richard}, \binits{G.}},
\bauthor{\bsnm{Serrurier}, \binits{M.}}:
\batitle{Analogical proportion-based methods for recommendation - first investigations}.
\bjtitle{FSS}
\bvolume{366},
\bfpage{110}--\blpage{132}
(\byear{2019})
\end{barticle}
\endbibitem

\bibitem{analogy-making-ai:2021:mitchel}
\begin{barticle}
\bauthor{\bsnm{Mitchell}, \binits{M.}}:
\batitle{Abstraction and analogy-making in artificial intelligence}.
\bjtitle{ANYAS}
\bvolume{1505},
\bfpage{79}--\blpage{101}
(\byear{2021})
\end{barticle}
\endbibitem

\bibitem{ap-functions-boolean-examples:2017:couceiro}
\begin{bchapter}
\bauthor{\bsnm{Couceiro}, \binits{M.}},
\bauthor{\bsnm{Hug}, \binits{N.}},
\bauthor{\bsnm{Prade}, \binits{H.}},
\bauthor{\bsnm{Richard}, \binits{G.}}:
\bctitle{Analogy-preserving functions: A way to extend boolean samples}.
In: \bbtitle{IJCAI},
pp. \bfpage{1575}--\blpage{1581}
(\byear{2017})
\end{bchapter}
\endbibitem

\bibitem{analogy-alea:2009:langlais}
\begin{bchapter}
\bauthor{\bsnm{Langlais}, \binits{P.}},
\bauthor{\bsnm{Yvon}, \binits{F.}},
\bauthor{\bsnm{Zweigenbaum}, \binits{P.}}:
\bctitle{Improvements in analogical learning: Application to translating multi-terms of the medical domain}.
In: \bbtitle{EACL},
pp. \bfpage{487}--\blpage{495}
(\byear{2009})
\end{bchapter}
\endbibitem

\bibitem{fam2016morphological}
\begin{bchapter}
\bauthor{\bsnm{Fam}, \binits{R.}},
\bauthor{\bsnm{Lepage}, \binits{Y.}}:
\bctitle{Morphological predictability of unseen words using computational analogy}.
In: \bbtitle{CAW@ICCBR},
vol. \bseriesno{1815},
pp. \bfpage{51}--\blpage{60}
(\byear{2016})
\end{bchapter}
\endbibitem

\bibitem{analogies-ml-perspective:2019:lim}
\begin{bchapter}
\bauthor{\bsnm{Lim}, \binits{S.}},
\bauthor{\bsnm{Prade}, \binits{H.}},
\bauthor{\bsnm{Richard}, \binits{G.}}:
\bctitle{Solving word analogies: {A} machine learning perspective}.
In: \bbtitle{ECSQARU},
vol. \bseriesno{11726},
pp. \bfpage{238}--\blpage{250}
(\byear{2019})
\end{bchapter}
\endbibitem

\bibitem{analogies-ml:2021:lim}
\begin{barticle}
\bauthor{\bsnm{Lim}, \binits{S.}},
\bauthor{\bsnm{Prade}, \binits{H.}},
\bauthor{\bsnm{Richard}, \binits{G.}}:
\batitle{Classifying and completing word analogies by machine learning}.
\bjtitle{IJAR}
\bvolume{132},
\bfpage{1}--\blpage{25}
(\byear{2021})
\end{barticle}
\endbibitem

\bibitem{minimal-complexity:2020:murena}
\begin{bchapter}
\bauthor{\bsnm{Murena}, \binits{P.-A.}},
\bauthor{\bsnm{Al-Ghossein}, \binits{M.}},
\bauthor{\bsnm{Dessalles}, \binits{J.-L.}},
\bauthor{\bsnm{Cornuéjols}, \binits{A.}}:
\bctitle{Solving analogies on words based on minimal complexity transformation}.
In: \bbtitle{IJCAI},
pp. \bfpage{1848}--\blpage{1854}
(\byear{2020})
\end{bchapter}
\endbibitem

\bibitem{detecting:2021:alsaidi}
\begin{bchapter}
\bauthor{\bsnm{Alsaidi}, \binits{S.}},
\bauthor{\bsnm{Decker}, \binits{A.}},
\bauthor{\bsnm{Lay}, \binits{P.}},
\bauthor{\bsnm{Marquer}, \binits{E.}},
\bauthor{\bsnm{Murena}, \binits{P.-A.}},
\bauthor{\bsnm{Couceiro}, \binits{M.}}:
\bctitle{A neural approach for detecting morphological analogies}.
In: \bbtitle{DSAA},
pp. \bfpage{1}--\blpage{10}
(\byear{2021})
\end{bchapter}
\endbibitem

\bibitem{transfer:2021:alsaidi}
\begin{bchapter}
\bauthor{\bsnm{Alsaidi}, \binits{S.}},
\bauthor{\bsnm{Decker}, \binits{A.}},
\bauthor{\bsnm{Lay}, \binits{P.}},
\bauthor{\bsnm{Marquer}, \binits{E.}},
\bauthor{\bsnm{Murena}, \binits{P.-A.}},
\bauthor{\bsnm{Couceiro}, \binits{M.}}:
\bctitle{{On the Transferability of Neural Models of Morphological Analogies}}.
In: \bbtitle{AIMLAI@ECML-PKDD},
vol. \bseriesno{1524},
pp. \bfpage{76}--\blpage{89}
(\byear{2021})
\end{bchapter}
\endbibitem

\bibitem{analogy-overview-extended-preprint:2021:alsaidi}
\begin{botherref}
\oauthor{\bsnm{Alsaidi}, \binits{S.}},
\oauthor{\bsnm{Decker}, \binits{A.}},
\oauthor{\bsnm{Marquer}, \binits{E.}},
\oauthor{\bsnm{Murena}, \binits{P.}},
\oauthor{\bsnm{Couceiro}, \binits{M.}}:
Tackling morphological analogies using deep learning - extended version.
CoRR
(2021)
\end{botherref}
\endbibitem

\bibitem{analogy-genmorpho:2022:chan}
\begin{bchapter}
\bauthor{\bsnm{Chan}, \binits{K.}},
\bauthor{\bsnm{Kaszefski-Yaschuk}, \binits{S.P.}},
\bauthor{\bsnm{Saran}, \binits{C.}},
\bauthor{\bsnm{Marquer}, \binits{E.}},
\bauthor{\bsnm{Couceiro}, \binits{M.}}:
\bctitle{{Solving Morphological Analogies Through Generation}}.
In: \bbtitle{IARML@IJCAI-ECAI},
vol. \bseriesno{3174},
pp. \bfpage{29}--\blpage{39}
(\byear{2022})
\end{bchapter}
\endbibitem

\bibitem{analogy-retrival:2022:marquer}
\begin{bchapter}
\bauthor{\bsnm{Marquer}, \binits{E.}},
\bauthor{\bsnm{Alsaidi}, \binits{S.}},
\bauthor{\bsnm{Decker}, \binits{A.}},
\bauthor{\bsnm{Murena}, \binits{P.-A.}},
\bauthor{\bsnm{Couceiro}, \binits{M.}}:
\bctitle{{A Deep Learning Approach to Solving Morphological Analogies}}.
In: \bbtitle{ICCBR}.
\bsertitle{LNCS},
vol. \bseriesno{13405},
pp. \bfpage{159}--\blpage{174}
(\byear{2022})
\end{bchapter}
\endbibitem

\bibitem{transfer:2022:marquer}
\begin{bchapter}
\bauthor{\bsnm{Marquer}, \binits{E.}},
\bauthor{\bsnm{Murena}, \binits{P.-A.}},
\bauthor{\bsnm{Couceiro}, \binits{M.}}:
\bctitle{{Transferring Learned Models of Morphological Analogy}}.
In: \bbtitle{ATA@ICCBR},
\bconflocation{Nancy, France}
(\byear{2022})
\end{bchapter}
\endbibitem

\bibitem{SadeghiZF15}
\begin{bchapter}
\bauthor{\bsnm{Sadeghi}, \binits{F.}},
\bauthor{\bsnm{Zitnick}, \binits{C.L.}},
\bauthor{\bsnm{Farhadi}, \binits{A.}}:
\bctitle{Visalogy: Answering visual analogy questions}.
In: \bbtitle{NeurIPS},
pp. \bfpage{1882}--\blpage{1890}
(\byear{2015})
\end{bchapter}
\endbibitem

\bibitem{PeyreSLS19}
\begin{bchapter}
\bauthor{\bsnm{Peyre}, \binits{J.}},
\bauthor{\bsnm{Laptev}, \binits{I.}},
\bauthor{\bsnm{Schmid}, \binits{C.}},
\bauthor{\bsnm{Sivic}, \binits{J.}}:
\bctitle{Detecting unseen visual relations using analogies}.
In: \bbtitle{ICCV},
pp. \bfpage{1981}--\blpage{1990}
(\byear{2019})
\end{bchapter}
\endbibitem

\bibitem{analogy-tsv:2022:zervakis}
\begin{bchapter}
\bauthor{\bsnm{Zervakis}, \binits{G.}},
\bauthor{\bsnm{Vincent}, \binits{E.}},
\bauthor{\bsnm{Couceiro}, \binits{M.}},
\bauthor{\bsnm{Schoenauer}, \binits{M.}},
\bauthor{\bsnm{Marquer}, \binits{E.}}:
\bctitle{{An analogy based approach for solving target sense verification}}.
In: \bbtitle{NLPIR},
\bconflocation{Bangkok, Thailand}
(\byear{2022})
\end{bchapter}
\endbibitem

\bibitem{copycat:1995:hofstadter-mitchel}
\begin{bchapter}
\bauthor{\bsnm{Hofstadter}, \binits{D.}},
\bauthor{\bsnm{Mitchell}, \binits{M.}}:
\bctitle{The copycat project: A model of mental fluidity and analogy-making}.
In: \bbtitle{FCCAs},
pp. \bfpage{205}--\blpage{267}
(\byear{1995}).
\bcomment{Chap. 5}
\end{bchapter}
\endbibitem

\bibitem{complexity-hofstadter:2017:murena}
\begin{bchapter}
\bauthor{\bsnm{Murena}, \binits{P.-A.}},
\bauthor{\bsnm{Dessalles}, \binits{J.-L.}},
\bauthor{\bsnm{Cornu{\'e}jols}, \binits{A.}}:
\bctitle{A complexity based approach for solving hofstadter's analogies}.
In: \bbtitle{CAW@ICCBR}
(\byear{2017})
\end{bchapter}
\endbibitem

\bibitem{tools-analogy:2018:fam-lepage}
\begin{bchapter}
\bauthor{\bsnm{Fam}, \binits{R.}},
\bauthor{\bsnm{Lepage}, \binits{Y.}}:
\bctitle{Tools for the production of analogical grids and a resource of n-gram analogical grids in 11 languages}.
In: \bbtitle{LREC},
pp. \bfpage{1060}--\blpage{1066}
(\byear{2018})
\end{bchapter}
\endbibitem

\bibitem{siganalogies}
\begin{botherref}
\oauthor{\bsnm{Marquer}, \binits{E.}},
\oauthor{\bsnm{Couceiro}, \binits{M.}},
\oauthor{\bsnm{Alsaidi}, \binits{S.}},
\oauthor{\bsnm{Decker}, \binits{A.}}:
Siganalogies - Morphological Analogies from Sigmorphon 2016 and 2019
\end{botherref}
\endbibitem

\bibitem{analogy-formal:2001:lepage}
\begin{bchapter}
\bauthor{\bsnm{Lepage}, \binits{Y.}}:
\bctitle{Analogy and formal languages},
vol. \bseriesno{53},
pp. \bfpage{180}--\blpage{191}
(\byear{2001})
\end{bchapter}
\endbibitem

\bibitem{analogical-dissimilarity:2008:miclet}
\begin{barticle}
\bauthor{\bsnm{Miclet}, \binits{L.}},
\bauthor{\bsnm{Bayoudh}, \binits{S.}},
\bauthor{\bsnm{Delhay}, \binits{A.}}:
\batitle{Analogical dissimilarity: Definition, algorithms and two experiments in machine learning}.
\bjtitle{JAIR}
\bvolume{32},
\bfpage{793}--\blpage{824}
(\byear{2008})
\end{barticle}
\endbibitem

\bibitem{ap:2022:antic}
\begin{barticle}
\bauthor{\bsnm{Antic}, \binits{C.}}:
\batitle{Analogical proportions}.
\bjtitle{AMAI}
\bvolume{90},
\bfpage{595}--\blpage{644}
(\byear{2022})
\end{barticle}
\endbibitem

\bibitem{analogy-between-concepts:2019:barbot-etal}
\begin{barticle}
\bauthor{\bsnm{Barbot}, \binits{N.}},
\bauthor{\bsnm{Miclet}, \binits{L.}},
\bauthor{\bsnm{Prade}, \binits{H.}}:
\batitle{{Analogy between concepts}}.
\bjtitle{AI}
\bvolume{275},
\bfpage{487}--\blpage{539}
(\byear{2019})
\end{barticle}
\endbibitem

\bibitem{lepage1996saussurian}
\begin{bchapter}
\bauthor{\bsnm{Lepage}, \binits{Y.}},
\bauthor{\bsnm{Ando}, \binits{S.}}:
\bctitle{Saussurian analogy: a theoretical account and its application}.
In: \bbtitle{COLING},
pp. \bfpage{717}--\blpage{722}
(\byear{1996})
\end{bchapter}
\endbibitem

\bibitem{trends-analogical-reasoning:2014:prade}
\begin{bchapter}
\bauthor{\bsnm{Prade}, \binits{H.}},
\bauthor{\bsnm{Richard}, \binits{G.}}:
\bctitle{A short introduction to computational trends in analogical reasoning}.
In: \bbtitle{CAAR-CT}.
\bsertitle{SCI},
vol. \bseriesno{548},
pp. \bfpage{1}--\blpage{12}
(\byear{2014})
\end{bchapter}
\endbibitem

\bibitem{analogy-commutation-linguistic-fr:2003:lepage}
\begin{botherref}
\oauthor{\bsnm{Lepage}, \binits{Y.}}:
{De l'analogie rendant compte de la commutation en linguistique}.
Habilitation {\`a} diriger des recherches,
{Universit{\'e} Joseph-Fourier - Grenoble I}
(2003)
\end{botherref}
\endbibitem

\bibitem{boolean-analogy:2018:couceiro}
\begin{bchapter}
\bauthor{\bsnm{Couceiro}, \binits{M.}},
\bauthor{\bsnm{Hug}, \binits{N.}},
\bauthor{\bsnm{Prade}, \binits{H.}},
\bauthor{\bsnm{Richard}, \binits{G.}}:
\bctitle{{Behavior of Analogical Inference w.r.t. Boolean Functions}}.
In: \bbtitle{IJCAI},
pp. \bfpage{2057}--\blpage{2063}
(\byear{2018})
\end{bchapter}
\endbibitem

\bibitem{galois-analogical-clf:2023:couceiro-lehtonen}
\begin{botherref}
\oauthor{\bsnm{Couceiro}, \binits{M.}},
\oauthor{\bsnm{Lehtonen}, \binits{E.}}:
{Galois theory for analogical classifiers}
(2023)
\end{botherref}
\endbibitem

\bibitem{glove:2014:pennington}
\begin{bchapter}
\bauthor{\bsnm{Pennington}, \binits{J.}},
\bauthor{\bsnm{Socher}, \binits{R.}},
\bauthor{\bsnm{Manning}, \binits{C.D.}}:
\bctitle{Glove: Global vectors for word representation}.
In: \bbtitle{EMNLP},
pp. \bfpage{1532}--\blpage{1543}
(\byear{2014})
\end{bchapter}
\endbibitem

\bibitem{finite-state-trancducers:2003:yvon}
\begin{botherref}
\oauthor{\bsnm{Yvon}, \binits{F.}}:
Finite-state transducers solving analogies on words.
Report GET/ENST\&LTCI
(2003)
\end{botherref}
\endbibitem

\bibitem{sigmorphon:2017:lepage}
\begin{bchapter}
\bauthor{\bsnm{Lepage}, \binits{Y.}}:
\bctitle{Character-position arithmetic for analogy questions between word forms}.
In: \bbtitle{CAW@ICCBR},
vol. \bseriesno{2028},
pp. \bfpage{23}--\blpage{32}
(\byear{2017})
\end{bchapter}
\endbibitem

\bibitem{vec-to-sec-sent-analogies:2020:lepage}
\begin{bchapter}
\bauthor{\bsnm{Wang}, \binits{L.}},
\bauthor{\bsnm{Lepage}, \binits{Y.}}:
\bctitle{Vector-to-sequence models for sentence analogies}.
In: \bbtitle{ICACSIS},
pp. \bfpage{441}--\blpage{446}
(\byear{2020})
\end{bchapter}
\endbibitem

\bibitem{morpho-thesis:2020:vania}
\begin{botherref}
\oauthor{\bsnm{Vania}, \binits{C.}}:
On understanding character-level models for representing morphology.
PhD thesis,
University of Edinburgh
(2020)
\end{botherref}
\endbibitem

\bibitem{3-cos-mul:2014:levy-goldberg}
\begin{bchapter}
\bauthor{\bsnm{Levy}, \binits{O.}},
\bauthor{\bsnm{Goldberg}, \binits{Y.}}:
\bctitle{Dependency-based word embeddings}.
In: \bbtitle{Short Papers}.
\bsertitle{ACL},
vol. \bseriesno{2},
pp. \bfpage{302}--\blpage{308}
(\byear{2014})
\end{bchapter}
\endbibitem

\bibitem{bert-multilingual:2019:delvin}
\begin{bchapter}
\bauthor{\bsnm{Devlin}, \binits{J.}},
\bauthor{\bsnm{Chang}, \binits{M.}},
\bauthor{\bsnm{Lee}, \binits{K.}},
\bauthor{\bsnm{Toutanova}, \binits{K.}}:
\bctitle{{BERT:} pre-training of deep bidirectional transformers for language understanding}.
In: \bbtitle{NAACL-HLT},
pp. \bfpage{4171}--\blpage{4186}
(\byear{2019})
\end{bchapter}
\endbibitem

\bibitem{efficient-representation-w2v:2013:mikolov}
\begin{bchapter}
\bauthor{\bsnm{Mikolov}, \binits{T.}},
\bauthor{\bsnm{Chen}, \binits{K.}},
\bauthor{\bsnm{Corrado}, \binits{G.}},
\bauthor{\bsnm{Dean}, \binits{J.}}:
\bctitle{Efficient estimation of word representations in vector space}.
In: \bbtitle{ICLR}
(\byear{2013})
\end{bchapter}
\endbibitem

\bibitem{fasttext:2017:bojanwoski}
\begin{barticle}
\bauthor{\bsnm{Bojanowski}, \binits{P.}},
\bauthor{\bsnm{Grave}, \binits{E.}},
\bauthor{\bsnm{Joulin}, \binits{A.}},
\bauthor{\bsnm{Mikolo}, \binits{T.}}:
\batitle{Enriching word vectors with subword information}.
\bjtitle{ACL}
\bvolume{5},
\bfpage{135}--\blpage{146}
(\byear{2017})
\end{barticle}
\endbibitem

\bibitem{cnn:1989:lecun}
\begin{barticle}
\bauthor{\bsnm{LeCun}, \binits{Y.}},
\bauthor{\bsnm{Boser}, \binits{B.E.}},
\bauthor{\bsnm{Denker}, \binits{J.S.}},
\bauthor{\bsnm{Henderson}, \binits{D.}},
\bauthor{\bsnm{Howard}, \binits{R.E.}},
\bauthor{\bsnm{Hubbard}, \binits{W.E.}},
\bauthor{\bsnm{Jackel}, \binits{L.D.}}:
\batitle{Backpropagation applied to handwritten zip code recognition}.
\bjtitle{NeurComp}
\bvolume{1}(\bissue{4}),
\bfpage{541}--\blpage{551}
(\byear{1989})
\end{barticle}
\endbibitem

\bibitem{ae:1991:kramer}
\begin{barticle}
\bauthor{\bsnm{Kramer}, \binits{M.A.}}:
\batitle{Nonlinear principal component analysis using autoassociative neural networks}.
\bjtitle{AIChE}
\bvolume{37}(\bissue{2}),
\bfpage{233}--\blpage{243}
(\byear{1991})
\end{barticle}
\endbibitem

\bibitem{lstm:1996:hochreiter}
\begin{bchapter}
\bauthor{\bsnm{Hochreiter}, \binits{S.}},
\bauthor{\bsnm{Schmidhuber}, \binits{J.}}:
\bctitle{Lstm can solve hard long time lag problems}.
In: \bbtitle{NeurIPS},
pp. \bfpage{473}--\blpage{479}
(\byear{1996})
\end{bchapter}
\endbibitem

\bibitem{fchollet-seq2seq}
\begin{botherref}
\oauthor{\bsnm{Chollet}, \binits{F.}}:
Character-level recurrent sequence-to-sequence model
(2017)
\end{botherref}
\endbibitem

\bibitem{bilstm:2005:graves}
\begin{barticle}
\bauthor{\bsnm{Graves}, \binits{A.}},
\bauthor{\bsnm{Schmidhuber}, \binits{J.}}:
\batitle{Framewise phoneme classification with bidirectional {LSTM} and other neural network architectures}.
\bjtitle{NN}
\bvolume{18}(\bissue{5-6}),
\bfpage{602}--\blpage{610}
(\byear{2005})
\end{barticle}
\endbibitem

\bibitem{linguistic-regularities:2013:mikolov}
\begin{bchapter}
\bauthor{\bsnm{Mikolov}, \binits{T.}},
\bauthor{\bsnm{Yih}, \binits{W.-T.}},
\bauthor{\bsnm{Zweig}, \binits{G.}}:
\bctitle{Linguistic regularities in continuous space word representations}.
In: \bbtitle{NAACL},
pp. \bfpage{746}--\blpage{751}
(\byear{2013})
\end{bchapter}
\endbibitem

\bibitem{wav2vec2:2020:develin}
\begin{bchapter}
\bauthor{\bsnm{Baevski}, \binits{A.}},
\bauthor{\bsnm{Zhou}, \binits{Y.}},
\bauthor{\bsnm{Mohamed}, \binits{A.}},
\bauthor{\bsnm{Auli}, \binits{M.}}:
\bctitle{wav2vec 2.0: {A} framework for self-supervised learning of speech representations}.
In: \bbtitle{NeurIPS},
pp. \bfpage{12449}--\blpage{12460}
(\byear{2020})
\end{bchapter}
\endbibitem

\bibitem{vit:2021:dosovitskiy-et-al}
\begin{bchapter}
\bauthor{\bsnm{Dosovitskiy}, \binits{A.}},
\bauthor{\bsnm{Beyer}, \binits{L.}},
\bauthor{\bsnm{Kolesnikov}, \binits{A.}},
\bauthor{\bsnm{Weissenborn}, \binits{D.}},
\bauthor{\bsnm{Zhai}, \binits{X.}},
\bauthor{\bsnm{Unterthiner}, \binits{T.}},
\bauthor{\bsnm{Dehghani}, \binits{M.}},
\bauthor{\bsnm{Minderer}, \binits{M.}},
\bauthor{\bsnm{Heigold}, \binits{G.}},
\bauthor{\bsnm{Gelly}, \binits{S.}},
\bauthor{\bsnm{Uszkoreit}, \binits{J.}},
\bauthor{\bsnm{Houlsby}, \binits{N.}}:
\bctitle{An image is worth 16x16 words: Transformers for image recognition at scale}.
In: \bbtitle{ICLR}
(\byear{2021})
\end{bchapter}
\endbibitem

\bibitem{sigmorphon:2016:cotterell}
\begin{bchapter}
\bauthor{\bsnm{Cotterell}, \binits{R.}},
\bauthor{\bsnm{Kirov}, \binits{C.}},
\bauthor{\bsnm{Sylak-Glassman}, \binits{J.}},
\bauthor{\bsnm{Yarowsky}, \binits{D.}},
\bauthor{\bsnm{Eisner}, \binits{J.}},
\bauthor{\bsnm{Hulden}, \binits{M.}}:
\bctitle{The sigmorphon 2016 shared task--morphological reinflection}.
In: \bbtitle{SIGMORPHON 2016},
pp. \bfpage{10}--\blpage{22}
(\byear{2016})
\end{bchapter}
\endbibitem

\bibitem{sigmorphon:2019:mccarthy-etal}
\begin{bchapter}
\bauthor{\bsnm{McCarthy}, \binits{A.D.}},
\bauthor{\bsnm{Vylomova}, \binits{E.}},
\bauthor{\bsnm{Wu}, \binits{S.}},
\bauthor{\bsnm{Malaviya}, \binits{C.}},
\bauthor{\bsnm{Wolf-Sonkin}, \binits{L.}},
\bauthor{\bsnm{Nicolai}, \binits{G.}},
\bauthor{\bsnm{Kirov}, \binits{C.}},
\bauthor{\bsnm{Silfverberg}, \binits{M.}},
\bauthor{\bsnm{Mielke}, \binits{S.J.}},
\bauthor{\bsnm{Heinz}, \binits{J.}},
\bauthor{\bsnm{Cotterell}, \binits{R.}},
\bauthor{\bsnm{Hulden}, \binits{M.}}:
\bctitle{The {SIGMORPHON} 2019 shared task: Morphological analysis in context and cross-lingual transfer for inflection}.
In: \bbtitle{CRPPM@ACL},
pp. \bfpage{229}--\blpage{244}
(\byear{2019})
\end{bchapter}
\endbibitem

\bibitem{jap-data:2018:karpinska}
\begin{bchapter}
\bauthor{\bsnm{Karpinska}, \binits{M.}},
\bauthor{\bsnm{Li}, \binits{B.}},
\bauthor{\bsnm{Rogers}, \binits{A.}},
\bauthor{\bsnm{Drozd}, \binits{A.}}:
\bctitle{Subcharacter information in japanese embeddings: when is it worth it?}
In: \bbtitle{RELNLP},
pp. \bfpage{28}--\blpage{37}
(\byear{2018})
\end{bchapter}
\endbibitem

\bibitem{morpho-portugese:1981:brakel}
\begin{barticle}
\bauthor{\bsnm{Brakel}, \binits{A.}}:
\batitle{Boundaries in a morphological grammar of portuguese}.
\bjtitle{WORD}
\bvolume{32}(\bissue{3}),
\bfpage{193}--\blpage{212}
(\byear{1981})
\end{barticle}
\endbibitem

\bibitem{neural:1994:haykin}
\begin{bbook}
\bauthor{\bsnm{Haykin}, \binits{S.}}:
\bbtitle{Neural Networks: a Comprehensive Foundation},
(\byear{1994})
\end{bbook}
\endbibitem

\bibitem{random-forest:1995:ho}
\begin{bchapter}
\bauthor{\bsnm{Ho}, \binits{T.K.}}:
\bctitle{Random decision forests}.
In: \bbtitle{ICDAR},
vol. \bseriesno{1},
pp. \bfpage{278}--\blpage{282}
(\byear{1995})
\end{bchapter}
\endbibitem

\bibitem{svm:1995:cortes}
\begin{barticle}
\bauthor{\bsnm{Cortes}, \binits{C.}},
\bauthor{\bsnm{Vapnik}, \binits{V.}}:
\batitle{Support-vector networks}.
\bjtitle{ML}
\bvolume{20}(\bissue{3}),
\bfpage{273}--\blpage{297}
(\byear{1995})
\end{barticle}
\endbibitem

\bibitem{medical-analogies:2022:alsaidi}
\begin{bchapter}
\bauthor{\bsnm{Alsaidi}, \binits{S.}},
\bauthor{\bsnm{Couceiro}, \binits{M.}},
\bauthor{\bsnm{Quennelle}, \binits{S.}},
\bauthor{\bsnm{Burgun}, \binits{A.}},
\bauthor{\bsnm{Garcelon}, \binits{N.}},
\bauthor{\bsnm{Coulet}, \binits{A.}}:
\bctitle{Exploring analogical inference in healthcare}.
In: \bbtitle{IARML@IJAI-ECAI}.
\bsertitle{CEUR-WP},
vol. \bseriesno{3174},
pp. \bfpage{40}--\blpage{50}
(\byear{2022})
\end{bchapter}
\endbibitem

\bibitem{adam:2014:kingma}
\begin{bchapter}
\bauthor{\bsnm{Kingma}, \binits{D.P.}},
\bauthor{\bsnm{Ba}, \binits{J.}}:
\bctitle{Adam: {A} method for stochastic optimization}.
In: \bbtitle{ICLR}
(\byear{2015})
\end{bchapter}
\endbibitem

\bibitem{nadam:2016:dozat}
\begin{bchapter}
\bauthor{\bsnm{Dozat}, \binits{T.}}:
\bctitle{Incorporating {Nesterov Momentum into Adam}}.
In: \bbtitle{ICLR},
pp. \bfpage{1}--\blpage{4}
(\byear{2016})
\end{bchapter}
\endbibitem

\end{thebibliography}
